\documentclass[10pt,twocolumn,letterpaper]{article}

\usepackage{tikz}
\usepackage{amssymb}
\usepackage[pagenumbers]{conference}

\usepackage{graphicx}
\usepackage{booktabs}
\usepackage{multirow}
\usepackage{colortbl}
\usepackage{amsmath,amsfonts}
\usepackage{comment}

\newsavebox{\tblbox}
\definecolor{Light}{rgb}{0.99, 0.92, 0.95}

\graphicspath{{../Figures/}}

\definecolor{conferenceblue}{rgb}{0.21,0.49,0.74}
\usepackage[pagebackref,breaklinks,colorlinks,allcolors=conferenceblue]{hyperref}

\title{MV-dVRK: A Multi-Viewpoint Benchmark for Spatial Surgical Perception
}

\author{
Guido Caccianiga$^{1,2}$,
Sergey Prokudin$^{2}$,
Yutong Chen$^{2}$,
Bernard Javot$^{1}$,
Rachael L'Orsa$^{1}$,\\
Omer Burak Aladağ$^{1}$,
Yarden Sharon$^{1}$,
Jens Rolinger$^{3}$,
Ivan Capobianco$^{4}$,
Anton Deguet$^{5}$,\\
Siyu Tang$^{2}$,
and Katherine J. Kuchenbecker$^{1}$ \\
\\
$^{1}$Max Planck Institute for Intelligent Systems. 
$^{2}$ETH Zürich.\\ 
$^{3}$Erbe Group. 
$^{4}$Tübingen University Hospital. 
$^{5}$Johns Hopkins University. \\
{\tt\small \{caccianiga@is.mpg.de kjk@is.mpg.de\}}
}

\begin{document}

\newcommand{\yes}{\checkmark}
\newcommand{\no}{\(\times\)}

\maketitle
\begin{abstract}

Large-scale training and refined optimization techniques have greatly improved sparse multi-view 3D reconstruction. Despite their relevance to surgery, such methods have never before been rigorously evaluated on real endoscopic images. Current clinical telerobots deploy a single stereo camera inside the patient, making multi-viewpoint data extremely rare. This paper presents \textbf{MV-dVRK}, the first ex-vivo surgical dataset to combine multiple exposure-synchronized stereo viewpoints with accurate surface geometry and camera poses. The static subset of the benchmark provides dense SfM reference geometry, validated against an industrial 3D scanner, together with ground-truth camera poses and sparse-view test sets. We use MV-dVRK to systematically compare zero-shot monocular, stereo, multi-stereo, and multi-view 3D reconstruction methods as the number of viewpoints increases. With two endoscopes, multi-stereo reconstruction achieves the highest coverage. With a third viewpoint, optimization-based multi-view methods perform best, covering 67\% of ground-truth surface points within a 1\,mm tolerance and recovering highly accurate relative camera poses. By contrast, feed-forward foundation models cover only 43\% of the ground-truth surface in the same setting. MV-dVRK also includes ten dynamic sequences spanning multiple surgical tasks, with increasing kinematic complexity and tissue deformation, providing a basis for future research in multi-viewpoint surgical perception. The project is available at: https://mv-dvrk.is.mpg.de.

\end{abstract}

\begin{figure}[t]
\centering
\includegraphics[width=\linewidth]{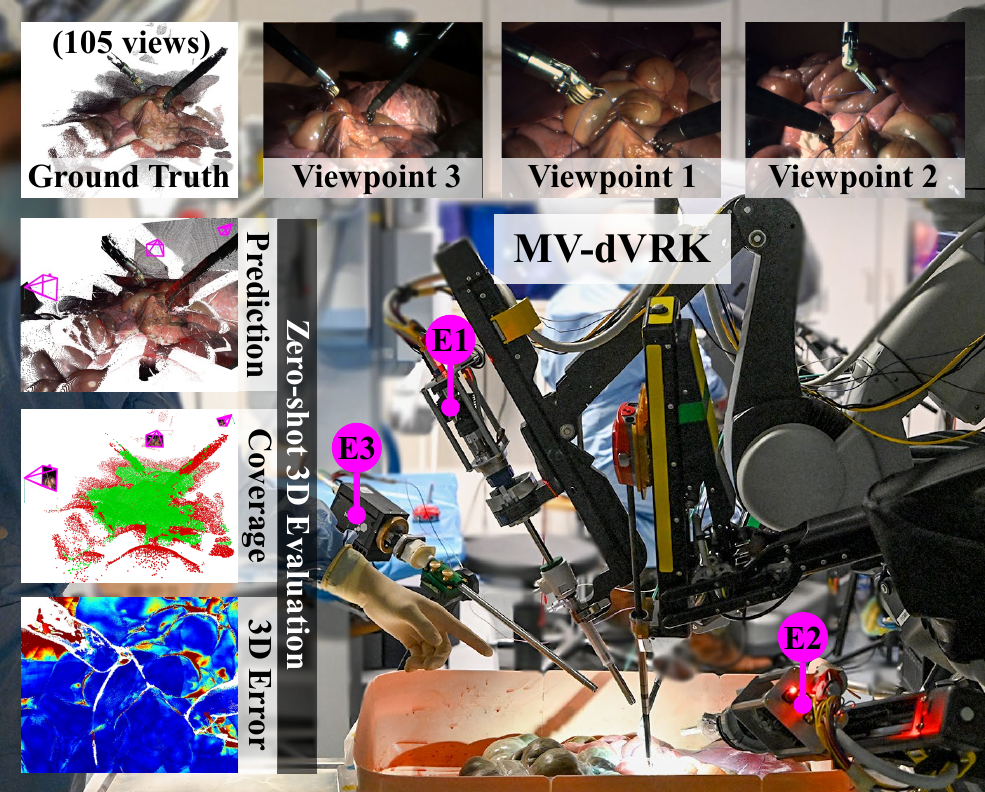}

\caption{Overview of the MV-dVRK benchmark workflow. Our custom robotic platform provides three exposure-synchronized stereo viewpoints, two of which are independently robot-controlled. For static scenes, dense multi-view SfM provides reference geometry and camera poses. Sparse zero-shot reconstructions from two to six synchronous input views are then evaluated for surface coverage, 3D error, and relative camera pose. The left inset shows a typical reconstruction, including the predicted 3D geometry, ground-truth surface coverage within a 1\,mm tolerance, and pointwise 3D error to the reference geometry. \\}
\label{fig:overview}
\end{figure}

\section{Introduction}
\label{sec:intro}

Accurate geometric reconstruction can improve robot-assisted minimally invasive surgery (RAMIS) by enabling advanced visualization for surgeons, such as free-viewpoint rendering~\cite{wang2022neural}, as well as geometry-aware robotic assistance~\cite{moccia2019vision} and autonomy~\cite{li2020super}. Today, however, surgical 3D perception is fundamentally constrained by the imaging setup: clinical telerobots provide only a single stereo viewpoint~\cite{boal2024evaluation}. Even sophisticated methods for handling tissue deformation and occlusions~\cite{song2026diff2dgs,yu2026endopairgs} cannot fully overcome the ambiguity of reconstructing unseen anatomy from a single viewpoint~\cite{zhu2024deformable}. Similarly, visuomotor policy learning benefits from additional viewpoints~\cite{moghani2025sufia}.

Multi-viewpoint imaging therefore offers a natural way to improve both geometric coverage and spatial awareness. At the same time, the physical constraints of minimally invasive surgery limit the number of cameras that can be deployed, making the relevant regime inherently sparse: only a few viewpoints, often with large baselines and limited overlap. This setting challenges modern 3D reconstruction methods and suffers from a scarcity of suitable surgical benchmarks. Existing datasets lack multiple exposure-synchronized endoscopic viewpoints and/or accurate measurements of scene geometry and camera poses (\S\ref{sec:relatedwork:multiview}).

We address this gap with \textbf{MV-dVRK}, a dataset collected using a multi-viewpoint surgical telerobot that we developed by extending the da Vinci Research Kit~\cite{kazanzides2014open} with a second independently controlled stereo endoscope, custom high-resolution cameras, and exposure-synchronized multi-camera acquisition (Fig.~\ref{fig:overview}, \S\ref{sec:platform}). For this benchmark, we add a fixed stereo endoscope, yielding three synchronized stereo viewpoints. The resulting data enable controlled evaluation of sparse multi-view 3D reconstruction on real ex-vivo surgical imagery and provide synchronized dynamic sequences for future research in multi-viewpoint surgical perception. Beyond multi-view deployment, the geometric reference also enables rigorous evaluation of methods designed to operate from a single endoscope.

This paper contributes:
\begin{itemize}
    \item \textbf{A multi-viewpoint surgical dataset} (\S\ref{sec:dataset}) containing both large-baseline static scenes with dense reference geometry and synchronized dynamic sequences spanning multiple surgical tasks. The static reference reconstructions provide tissue-surface geometry and camera poses and are independently validated against an industrial 3D scanner.

    \item \textbf{A sparse-view 3D reconstruction benchmark} (\S\ref{sec:evaluation}) for systematically comparing monocular, stereo, multi-stereo, and multi-view methods as the number of synchronized input viewpoints increases.

    \item \textbf{An analysis of current multi-view 3D perception methods} (\S\ref{sec:results}), showing how reconstruction coverage, geometric accuracy, and camera-pose estimation change with viewpoint count and view overlap. The results reveal complementary strengths of stereo and jointly optimized multi-view reconstruction and identify promising directions for practical multi-view surgical perception.
\end{itemize}
\section{Related work}
\label{sec:relatedwork}

\subsection{Multi-view surgical datasets}
\label{sec:relatedwork:multiview}

Most existing datasets for robotic surgery provide only a single viewpoint and face a fundamental trade-off between realistic data and accurate geometric ground truth. Measuring deforming tissue with sub-millimeter accuracy remains challenging both ex vivo and intraoperatively. Surgical simulators provide exact reference geometry and controlled variation but do not fully reproduce the visual appearance and tissue mechanics of real surgery. When available for real data, shape ground truth is typically obtained using a scanner, structured light, or computed tomography~\cite{penza2018endoabs,allan2021stereo,edwards2022serv}; the cost, time, and disruption of these approaches limit the scale and diversity of the resulting datasets. Larger in-vivo collections generally provide no dense reference geometry of the tissue surface~\cite{hayoz2023learning,zia2025surgical,jiang2025surgisr4k}.

Only recently have surgical datasets begun to include multiple viewpoints. ImitateCholec~\cite{hansen2026imitatecholec} supplements a stationary stereo endoscope with two wrist-mounted cameras, but the streams are software-synchronized, limiting their use for precise multi-view geometric evaluation under deformation. Concurrent work, iMED~\cite{bonilla2026imed}, introduces a large dual-stereo-endoscope dataset spanning ex-vivo specimens, human cadavers, and an in-vivo porcine model. One endoscope is robot-mounted on a da Vinci~5 arm, while the second provides a fixed nearby perspective (inter-viewpoint baseline $\leq$\,4\,cm\,,\,$\leq$\,30\,deg). iMED supports camera-pose evaluation via ArUco~\cite{garrido-jurado2014automatic} markers but does not provide dense reference geometry of the tissue surface. No existing dataset combines multiple robot-controlled viewpoints, large baselines, sensor exposure synchronization, accurate reference geometry, and camera poses (Tab.~\ref{tab:dataset_comparison}). MV-dVRK is designed to fill this gap.

\begin{table*}[t]

\caption{Comparison with existing surgical datasets. Among the datasets considered, MV-dVRK uniquely combines multiple independently moving viewpoints, exposure-synchronized acquisition, and accurate reference geometry and camera poses. The static subset (I) provides geometric and pose ground truth, while the dynamic subset (II) captures synchronized tissue deformation and camera motion.}
\label{tab:dataset_comparison}

\centering
\footnotesize
\renewcommand{\arraystretch}{1.2}
\newcommand{\maybe}{\ensuremath{\approx}}

\resizebox{\linewidth}{!}{%
\begin{tabular}{@{}llccccccccccc@{}}
\hline
\textbf{Dataset} &
\textbf{Yr.} &
\textbf{Type} &
\textbf{Origin} &
\textbf{VP} &
\textbf{Defor.} &
\textbf{Mov.\ Cam.} &
\textbf{Sync} &
\textbf{Geom.\ GT} &
\textbf{Pose GT} &
\textbf{Size} &
\textbf{Hz} &
\textbf{Resolution} \\
\hline

Hamlyn~\cite{hamlyn}
& '10 & In-vivo & Pig/Human & 1
& \yes & \yes & \no
& \no & \no
& $>10^5$ & 30 & 720\,$\times$\,288 \\

SCARED~\cite{allan2021stereo}
& '21 & Ex-vivo & Pig & 1
& \no & \yes & \no
& \yes~Struc.\ light & \maybe~Robot FK~(1\,VP)
& 40 & 0 & 1280\,$\times$\,720\phantom{0} \\

SERV-CT~\cite{edwards2022serv}
& '22 & Ex-vivo & Pig & 1
& \no & \no & \no
& \yes~CT & \yes~Manual
& 16 & 0 & 720\,$\times$\,576 \\

EndoNeRF~\cite{wang2022neural}
& '22 & In-vivo & Pig & 1
& \yes & \no & \no
& \no & \no
& 807 & 15 & 640\,$\times$\,512 \\

StereoMIS~\cite{hayoz2023learning}
& '23 & In-vivo & Pig & 1
& \yes & \yes & \no
& \no & \maybe~Robot FK~(1\,VP)
& $>10^4$ & 30 & 1280\,$\times$\,1024 \\

SurgVU~\cite{zia2025surgical}
& '25 & In-vivo & Pig & 1
& \yes & \yes & \no
& \no & \no
& $>10^6$ & 60 & 1280\,$\times$\,720\phantom{0} \\

SurgiSR4K~\cite{jiang2025surgisr4k}
& '25 & In-vivo & Pig & 1
& \yes & \yes & \no
& \no & \no
& $>10^3$ & 60 & 3840\,$\times$\,2160 \\

ImitateCholec~\cite{hansen2026imitatecholec}
& '26 & Ex-vivo & Pig & 3
& \yes & \no & \maybe~Software 
&\no & \maybe~Robot FK~(1\,VP)
& $>10^4$ & 30 & 960\,$\times$\,540 \\

iMED~\cite{bonilla2026imed}
& '26 & Ex-/In-vivo & Pig/Human & 2
& \yes & \yes & \maybe~Capture
& \no & \yes~ArUco~(2\,VP)
& $>10^5$ & 60 & 1280\,$\times$\,1024 \\

\hline

\textbf{MV-dVRK I (ours)}
& '26 & Ex-vivo & Pig & 3
& \no & \yes & \yes~Exposure
& \yes~Dense SfM & \yes~Dense SfM~(3\,VP)
& 840 & 0 & 2560\,$\times$\,2048 \\

\textbf{MV-dVRK II (ours)}
& '26 & Ex-vivo & Pig & 3
& \yes & \yes & \yes~Exposure
& \no & \yes~ArUco\,+\,Robot FK~(3\,VP)
& $>10^4$ & 30 & 2560\,$\times$\,2048 \\

\hline
\end{tabular}%
}

\vspace{1mm}
\scriptsize
\raggedright
VP = viewpoints;
Defor.\ = deforming tissue;
Mov.\ Cam.\ = moving endoscopic camera;
Sync = synchronization type: Software = timestamped by operating system; Capture = timestamped by capture card; Exposure = timestamped by image sensors. 
Geom.\ GT = ground-truth surface geometry;
CT = computed tomography;
Pose GT = ground-truth camera poses;
FK = forward kinematics;
Size = number of frames.
\yes, \maybe, and \no indicate availability. 
Pose GT cells also report the number of viewpoints tracked. 

\end{table*}

\subsection{Geometric reconstruction in computer and medical vision}
\label{sec:relatedwork:cv}

\textbf{Stereo reconstruction.}
Stereo correspondence~\cite{wen2025foundationstereo,jing2023uncertainty,lipson2021raft,wang2024selective} provides a direct route to metric depth from a calibrated camera pair and is particularly relevant to robot-assisted surgery, where current systems already deploy a stereo endoscope. Surgical stereo reconstruction has therefore been widely studied for recovering tissue geometry from a single endoscopic viewpoint~\cite{stoyanov2010realtime}. Recent learned approaches have further improved robustness and generalization~\cite{cheng2022deep}. In our benchmark, this family of approaches is represented by FoundationStereo~\cite{wen2025foundationstereo}, a recent method designed for zero-shot stereo matching. Stereo can provide accurate local geometry where the tissue is texture-rich and well-illuminated, but a single endoscopic viewpoint remains limited by occlusions and incomplete scene coverage. 

\textbf{Single-view surgical reconstruction.}
Because surgeons typically perform laparoscopic surgery with only a single camera view, much of the surgical reconstruction literature has focused on recovering geometry from monocular video~\cite{mahmoud2019live}. Recent methods have also adapted learned 3D priors to endoscopic imagery~\cite{sheikh2025endo,guo2025endo3r}. We select Endo3R~\cite{guo2025endo3r} as our monocular baseline because it is designed for online endoscopic reconstruction and can accumulate geometry over time. This choice allows us to test whether temporal observations from a single moving camera can achieve the coverage provided by additional simultaneous viewpoints. Nevertheless, reconstructing deforming anatomy (seen and particularly unseen) from a single camera remains fundamentally ill-posed~\cite{tretschk2023state,schmidt2024tracking}.

\textbf{Feed-forward multi-view reconstruction.}
Recent feed-forward (FF) foundation models significantly improve sparse multi-view 3D reconstruction. DUSt3R~\cite{wang2024dust3r} can directly regress camera parameters and dense 3D point maps from uncalibrated images, inspiring a rapidly growing family of models for sparse multi-view geometric prediction~\cite{leroy2024mast3r,wang2025vggt,wang2025pi3x,lin2025depth}. The ability to reconstruct from very sparse unposed input views makes these methods particularly attractive for surgical reconstruction. However, how these foundation models perform on endoscopic imagery, which is under-represented in their training data, requires further investigation. MV-dVRK provides an extensive benchmark for this purpose; we evaluate VGGT~\cite{wang2025vggt}, Pi3~\cite{wang2025pi}, Pi3X~\cite{wang2025pi3x}, and Depth Anything~3~\cite{lin2025depth}.

\textbf{Geometry-refined multi-view reconstruction.}
Structure from motion (SfM) provides the classical framework for jointly recovering camera poses and 3D structure from overlapping images~\cite{hartley2003mvg}, with COLMAP~\cite{schoenberger2016sfm} serving as a standard implementation. Classical feature-based SfM is most reliable with many overlapping observations of a static scene, whereas minimally invasive surgery provides only a few viewpoints with potentially large viewpoint change and limited overlap. Recent methods tackle this sparse regime by combining learned correspondence with explicit geometric optimization~\cite{wang2024vggsfm,duisterhof2025mast3r,chen2026ggpt}. We evaluate MASt3R-SfM~\cite{duisterhof2025mast3r} and GGPT~\cite{chen2026ggpt} as representative methods from this family. The former jointly optimizes camera poses and geometry from learned correspondences, while the latter consists of an SfM stage and a point-map refinement stage. Evaluating GGPT with several feed-forward backbones further allows us to directly compare feed-forward inference with geometry-refined multi-view reconstruction under the same sparse surgical inputs.

\begin{figure*}[t]
\centering
\includegraphics[width=\linewidth]{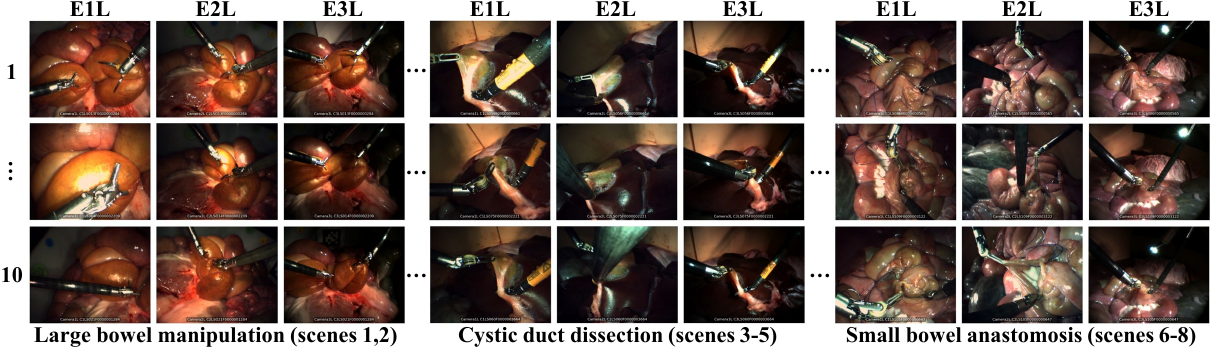}
\caption{MV-dVRK I test set. The three endoscopic viewpoints (E1, E2, E3) span different surgical tasks, tissue appearances, relative camera poses, and robot configurations; ``L'' denotes the left camera of each stereo endoscope. The scenes include large linear and angular baselines, challenging illumination, and regions that are occluded from one viewpoint but visible from another. Rows index the 10 selected synchronized keyframes per scene; intermediate keyframes are omitted for visualization.}
\label{fig:tiledImages}
\end{figure*}

\section{The MV-dVRK dataset}
\label{sec:dataset}

\subsection{Data acquisition platform}
\label{sec:platform}

The acquisition platform extends the da Vinci Research Kit (dVRK)~\cite{kazanzides2014open} with a second stereo endoscope moved by an independent robotic arm. We refer to this multi-viewpoint extension as the MV-dVRK platform. For the benchmark, we add a stereo endoscope on a fixed mount, yielding three synchronized stereo viewpoints (Fig.~\ref{fig:overview}).

The endoscopes are custom-built from surgical-grade optics and industrial global-shutter cameras. A shared hardware trigger exposes all six imaging sensors simultaneously, and their clocks are aligned using the Precision Time Protocol (PTP~\cite{ieee1588}), resulting in microsecond-level timestamp agreement across cameras. This exposure synchronization ensures that all cameras observe the same spatiotemporal state for a given frame, which is essential for geometric evaluation of deforming scenes. Intrinsic and stereo calibration (board: ChArUco DICT\textunderscore5$\times$5, 15$\times$24, 150\,mm$\times$100\,mm, ceramic lithography) achieve sub-pixel reprojection errors, and robot kinematics are logged at 100\,Hz on the same PTP network.

\subsection{Dataset collection and composition}

We recorded MV-dVRK during mock surgery on the freshly excised abdominal organs of two adult pigs that were sacrificed for other reasons by a local meat producer. Two expert visceral surgeons teleoperated various left and right instruments to perform representative surgical tasks, including cholecystectomy, tubular dissection, and small-bowel anastomosis. At selected stages of each procedure, we locked the instruments and swept each robotic endoscope over the stationary tissue to acquire dense geometric reference data. The recordings form two complementary subsets, summarized relative to existing surgical datasets in Tab.~\ref{tab:dataset_comparison}.

\textbf{MV-dVRK I (static).}
The static subset comprises eight geometric reference scenes spanning three surgical tasks and different tissue appearances, robot configurations, and camera geometries, as illustrated in Fig.~\ref{fig:tiledImages}. Each scene is acquired while the anatomy is stationary, allowing dense multi-view SfM to recover reference tissue-surface geometry and camera poses. From each scene, we select 10 synchronized keyframes to maximize diversity in relative camera poses, yielding 80 test instances with three stereo viewpoints each. Because all keyframes within a scene share the same underlying anatomy and reference reconstruction, our analysis treats the scene as the statistical unit, rather than the individual frame (\S\ref{sec:results:stats}). MV-dVRK I forms the quantitative core of this paper and provides the reference geometry, camera poses, and synchronized sparse inputs used throughout the benchmark.

\textbf{MV-dVRK II (dynamic).}
The dynamic subset contains ten long multi-view sequences recorded during active surgical manipulation, together with synchronized dVRK kinematics and per-scene calibration offsets used to link the robot measurements to the three endoscopes' optical frames, effectively delivering a continuous reference for both single- and multi-camera pose estimation. Unlike conventional single-endoscope recordings, these sequences capture substantial tool--tissue interaction and tissue deformation from both static and moving viewpoints simultaneously. This subset thus provides a complementary dynamic counterpart to the static subset of the benchmark and extends MV-dVRK to settings involving camera motion and non-rigid scene dynamics. Representative sequences appear in the supplementary video. These synchronized multi-view recordings were mainly designed to support future work on dynamic surgical 3D perception; they also provide out-of-the-box multi-view input for tasks such as dynamic novel-view synthesis and camera-pose estimation.

\section{Benchmark design}
\label{sec:evaluation}

MV-dVRK I provides a common protocol for evaluating sparse 3D reconstruction on real surgical imagery. Each test instance contains three synchronized stereo viewpoints, corresponding to six images in total. We evaluate subsets of these inputs while varying the number of viewpoints from one to three as the primary experimental variable. Depending on the reconstruction family, a viewpoint may contribute either one monocular image or a calibrated stereo pair. For every prediction, we use the same dense reference reconstruction (\S\ref{sec:evaluation:gt}), sparse test inputs (\S\ref{sec:evaluation:testset}), registration protocol (\S\ref{sec:evaluation:registration}), and evaluation metrics (\S\ref{sec:evaluation:metrics}).

\subsection{Dense reference reconstruction}
\label{sec:evaluation:gt}

\textbf{Rationale.}
The benchmark requires both dense tissue-surface geometry and camera poses in a common coordinate frame. External scanners can provide accurate surface geometry but do not simultaneously recover the endoscopic camera poses and are difficult to deploy during surgical acquisition~\cite{edwards2022serv,caccianiga2024dense}. We therefore construct the benchmark reference directly from dense multi-view SfM over the static acquisition sequences and use an industrial 3D scanner for independent geometric validation and metric scaling.

The reference and benchmark inputs originate from the same acquisitions of static scenes but operate in very different regimes. Each reference reconstruction uses a dense set of views, whereas evaluated methods receive only sparse synchronized images. This design provides a reference pose for every test image and, because test images are included in the reference reconstruction, enables pixel-aligned comparison between predicted and reference 3D points.

\textbf{Reference image selection.}
We subsample the three left-camera streams (E1L, E2L, E3L) from each static acquisition at 0.4\,Hz and manually remove blurry and redundant frames. 105 images per scene were retained for reference reconstruction. The third endoscope is sampled less densely because repeated frames from this stationary camera provide little additional viewpoint diversity for SfM.

\textbf{Dense reconstruction.}
We construct the reference using the SfM stage of GGPT~\cite{chen2026ggpt}. The pipeline uses cameras and points predicted by VGGT~\cite{wang2025vggt} as initialization and adopts RoMaV2~\cite{edstedt2025roma} to extract dense correspondences. It runs a global bundle adjustment using a compact set of high-confidence correspondences to optimize camera poses, upon which 3D points are triangulated from a dense set of correspondences. Uncertain and inconsistent points are filtered based on their triangulation angles and reprojection errors. Compared with previous SfM pipelines operating on sparse keypoint correspondences~\cite{wang2024vggsfm,schoenberger2016sfm}, GGPT-SfM is more robust to the low texture and specularities common in endoscopic imagery and provides substantially denser geometric support on our acquisitions. While classical SfM reconstructs only a few thousand points in this setting, the resulting reference models contain more than 14 million points per scene. Note that although we include GGPT in our benchmark, our reference construction uses only its SfM stage; the second stage of feed-forward refinement is not applied here, as it is not strictly geometry-grounded.

\begin{figure}[t]
\centering
\includegraphics[width=\linewidth]{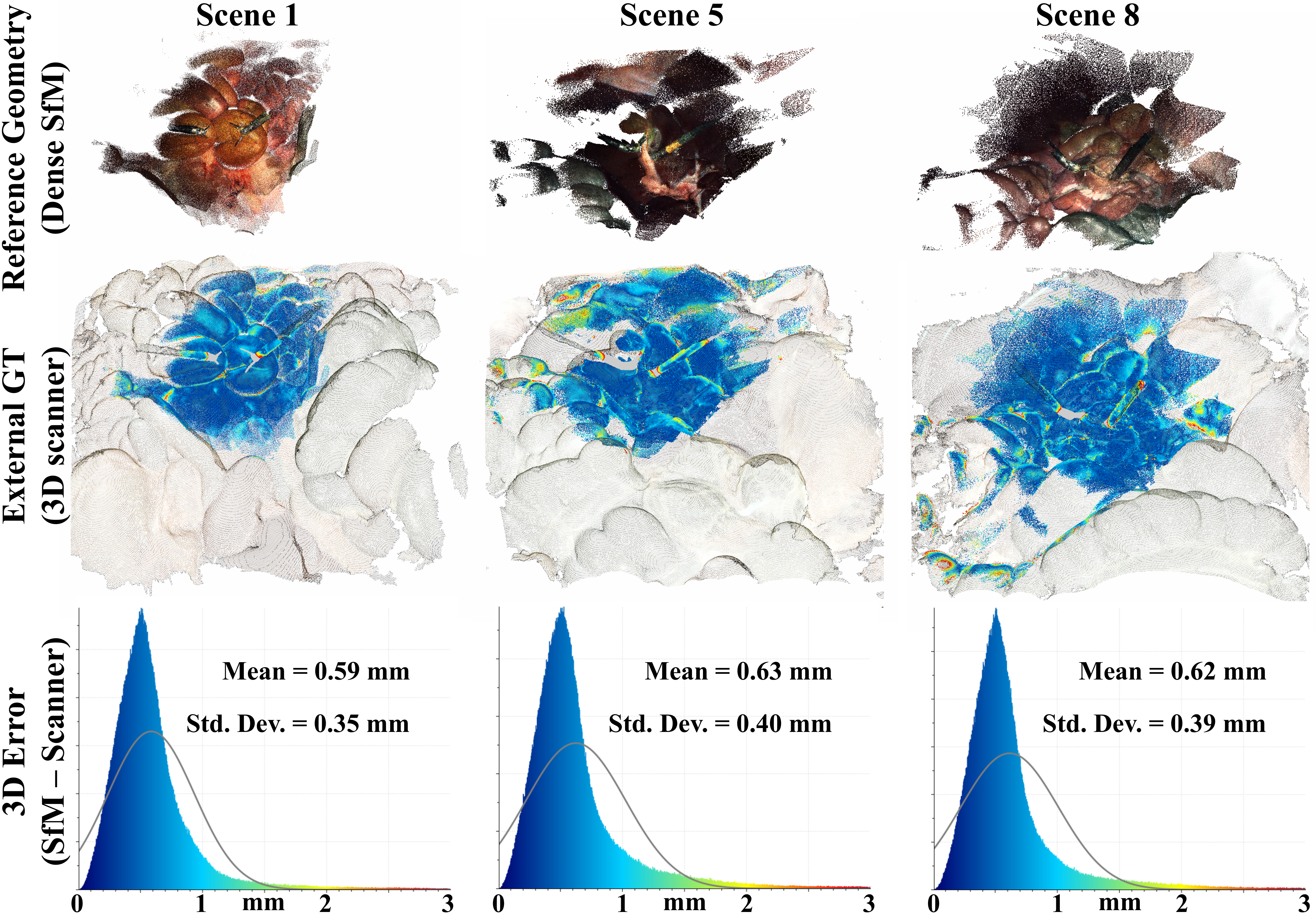}
\caption{External validation of the dense SfM reference geometry. For three representative scenes, we compare the SfM reconstruction (top) with an independently acquired 3D scanner surface (middle) and visualize the resulting point-to-surface error distribution (bottom). Mean errors are about 0.6\,mm across these scenes, supporting sub-millimeter accuracy of the reference geometry.}
\label{fig:GT_validation}
\end{figure}

\textbf{External validation.}
An industrial 3D scanner (Artec Eva~\cite{arteceva}) independently records the tissue surface once per surgical session. The scanner does not provide endoscopic camera poses and is therefore used for validation and metric scale rather than as the benchmark reference itself. Scaled ICP alignment~\cite{inproceedings} is done in CloudCompare v2.12. The mean point-to-surface distance (Fig.~\ref{fig:GT_validation}) between the dense SfM reconstructions and scanner meshes is $0.6 \pm 0.4$\,mm.  

\subsection{Synchronized sparse test set}
\label{sec:evaluation:testset}

For each of the eight reference scenes, we select 10 synchronized test instants to maximize diversity in relative camera poses, yielding 80 test samples. Each sample contains three stereo pairs, i.e., six synchronized images. The selected configurations have widely varying relative endoscope poses, amounts of view overlap, illumination conditions, robot configurations, and tool poses (Fig.~\ref{fig:tiledImages}).

The dense surface reconstruction provides projected reference geometry for every test image. Coverage differs across the three endoscopes because their fields of view observe different portions of the tissue. E1L has the highest reference completeness and is therefore used as the common reference view for pixel-aligned 3D error. The benchmark evaluates different subsets of the synchronized inputs while varying the number of available viewpoints.

\begin{figure*}[t!]
\centering
\includegraphics[width=\linewidth]{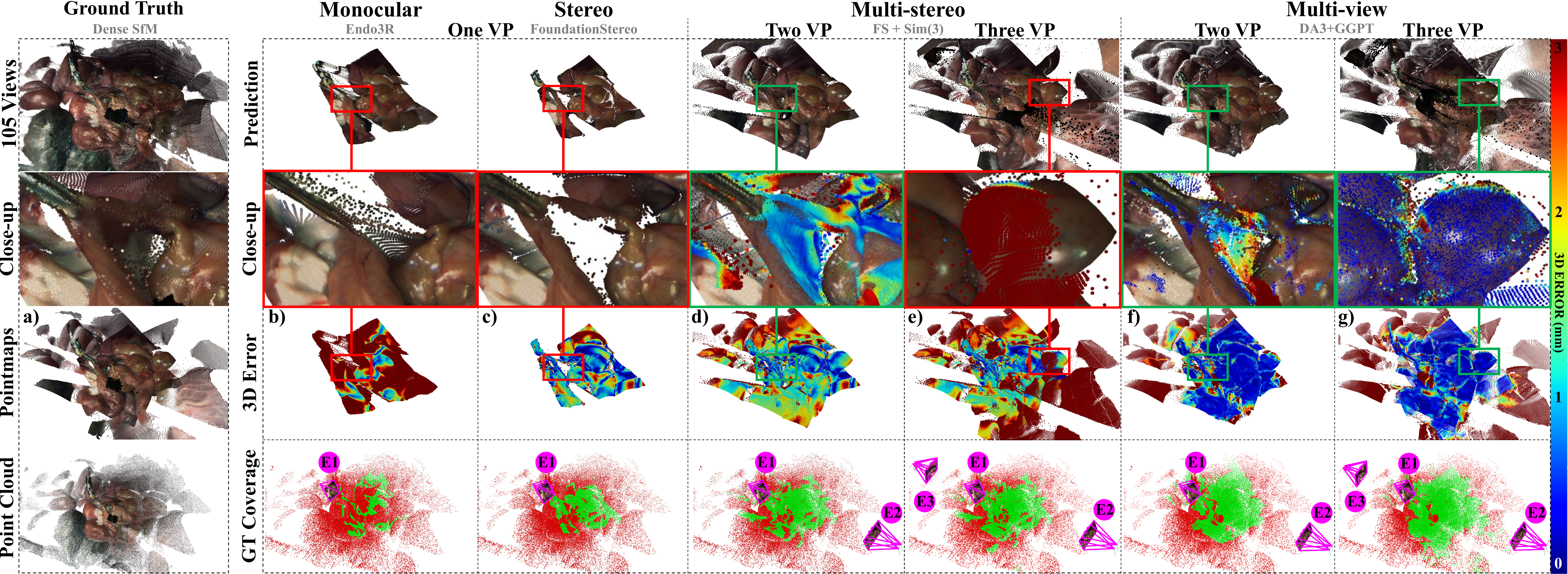}

\caption{Qualitative comparison of reconstruction paradigms as viewpoint count increases. 
(a) Dense SfM reference geometry, shown as the E1L pointmap used for pixel-aligned 3D error and as the full point cloud used for coverage evaluation. For each prediction, the rows show the reconstructed geometry, a close-up, pointwise 3D error, and ground-truth surface coverage within 1\,mm. 
With a single viewpoint, occluded anatomy is either incorrectly completed by Endo3R~\cite{guo2025endo3r} (b) or remains unobserved with FoundationStereo~\cite{wen2025foundationstereo} (c). 
Adding stereo viewpoints recovers previously occluded surface regions (d), but independent reconstruction and registration can introduce duplicated structures and large local errors (e). 
Joint multi-view reconstruction with DA3+GGPT~\cite{chen2026ggpt} produces more consistent geometry (f,g), with the third viewpoint substantially reducing error in the common reference region (g).}
\label{fig:registration_offset}
\end{figure*}

\subsection{Prediction-to-reference registration}
\label{sec:evaluation:registration}

Methods may express their reconstructions in different coordinate systems and, in some cases, only up to an unknown global scale. We therefore align every prediction to the reference before evaluating geometry. The benchmark measures reconstructed shape and relative camera geometry after alignment; recovery of absolute metric scale is not itself an evaluation target.

Because each test image is included in the dense reference reconstruction, predicted and reference points corresponding to the same E1L pixels provide direct 3D correspondences. We estimate a robust Sim(3) transformation from these correspondences using PyCOLMAP's \texttt{estimate\_sim3d\_robust} function~\cite{schoenberger2016sfm}, which combines LO-RANSAC~\cite{LORANSAC_DBLP:conf/dagm/ChumMK03} with Umeyama fitting~\cite{Umeyama88573}. We call this full point-based similarity alignment \emph{fitted} registration.

For methods that jointly predict geometry and camera poses, we additionally consider a \emph{camera-anchored Sim(3)} registration. Here, the predicted E1L camera determines the rotation and origin of the alignment, while only the global scale $s$ is estimated from the robust 3D inliers. Unlike a fully fitted Sim(3), this registration cannot independently rotate or translate the reconstructed geometry to compensate for inconsistencies with the method's own camera estimate. The resulting transformation is applied consistently to all predicted geometry and camera poses. Methods without a directly comparable reference-camera pose are evaluated using fitted registration, as specified in \S\ref{sec:implementation}.

\subsection{Metrics}
\label{sec:evaluation:metrics}

We evaluate the following three complementary properties for each reconstruction; formal metric definitions are provided in the supplementary material.

\textbf{Ground-truth coverage (Cov)} measures the fraction of reference surface points that have a predicted neighbor within a fixed distance threshold. Coverage within 1\,mm is our primary measure, and within 5\,mm is secondary. Point clouds are subsampled to $10^5$ points before evaluation.

\textbf{Mean 3D error (ME)} measures the Euclidean distance between registered predicted points and their pixel-aligned reference counterparts in the E1L field of view. E1L is shared across all benchmark configurations and therefore provides a common region for judging geometric accuracy.

\textbf{Relative pose error (Pose)} measures the accuracy of predicted relative transforms between the left cameras of different endoscopes. Translation error is reported in millimeters and geodesic rotation error in degrees.

\subsection{Statistical analysis}
\label{sec:results:stats}

The eight scenes are the units of statistical analysis. The ten test samples within each scene share the same anatomy and reference reconstruction and are therefore not treated as independent observations. For each pairwise comparison between methods, we first average the metric within each scene and then compute the eight paired scene-level differences, $\Delta$. We report their mean, the corresponding 95\% confidence interval ($t_7$), and the number of scenes in which the difference has the same sign ($k/8$). Result tables report the mean over all 80 test samples for compact descriptive comparison, while all inferential statements in the text are based on the paired scene-level analysis. We do not claim that the observed effect magnitudes generalize to new specimens, procedure types, or in-vivo conditions. Correlations across scenes are reported using Pearson's $r$ and interpreted descriptively given the small number of scenes.

\section{Benchmark analysis}
\label{sec:results}

\begin{table}[b]
\centering\scriptsize
\setlength{\fboxsep}{1.5pt}
\setlength{\tabcolsep}{0.5pt}\renewcommand{\arraystretch}{1.25}
\caption{Comparison of 3D reconstruction paradigms across viewpoint count. Registered multi-stereo is most accurate in the sparse two-viewpoint setting, while a third viewpoint favors jointly optimized multi-view reconstruction: Depth Anything~3~\cite{lin2025depth} + GGPT~\cite{chen2026ggpt} achieves the highest coverage (Cov) and lowest mean error (ME) in E1L, whereas FoundationStereo~\cite{wen2025foundationstereo} + Sim(3) obtains the lowest error in E2L. Darker shading indicates better performance; outlined cells denote the best value in each column.}

\label{tab:results_approaches_compared}
\begin{tabular}{l l cc c cc cc}
\addlinespace[2pt]
\textbf{Family} & \textbf{Method} & \shortstack{\rule{0pt}{6pt}{\tiny \textbf{View-}}\\{\rule[-1.6pt]{0pt}{5pt}\tiny \textbf{points}}} & \shortstack{\rule{0pt}{6pt}{\tiny \textbf{Sync.}}\\{\rule[-1.6pt]{0pt}{5pt}\tiny \textbf{views}}} & \shortstack{\rule{0pt}{6pt}{\tiny \textsf{\textbf{Cov}}\,$\uparrow$}\\{\rule[-1.6pt]{0pt}{5pt}\tiny \%@\textbf{1mm}}} & \shortstack{\rule{0pt}{6pt}{\tiny \textsf{\textbf{ME}}\,$\downarrow$}\\{\rule[-1.6pt]{0pt}{5pt}\tiny mm @\textbf{E1L}}} & \shortstack{\rule{0pt}{6pt}{\tiny \textsf{\textbf{ME}}\,$\downarrow$}\\{\rule[-1.6pt]{0pt}{5pt}\tiny mm @\textbf{E2L}}} & \shortstack{\rule{0pt}{6pt}{\tiny \textsf{\textbf{Pose t}}\,$\downarrow$}\\{\rule[-1.6pt]{0pt}{5pt}\tiny mm}} & \shortstack{\rule{0pt}{6pt}{\tiny \textsf{\textbf{Pose r}}\,$\downarrow$}\\{\rule[-1.6pt]{0pt}{5pt}\tiny deg}} \\
\midrule
\multirow{2}{*}{\textit{Monocular}} & Endo3R~\cite{guo2025endo3r} {\tiny (2 fr.)} & 1 & 1 & \colorbox[HTML]{DFFDE4}{\makebox[0.46cm][c]{\textcolor[HTML]{000000}{\tiny 12}}} & \colorbox[HTML]{FFEBDA}{\makebox[0.58cm][c]{\textcolor[HTML]{000000}{\tiny 17.4}}} & \makebox[0.58cm][c]{\tiny ---} & \makebox[0.52cm][c]{\tiny ---} & \makebox[0.46cm][c]{\tiny ---} \\
 & Endo3R~\cite{guo2025endo3r} {\tiny (50 fr.)} & 1 & 1 & \colorbox[HTML]{A2D0AA}{\makebox[0.46cm][c]{\textcolor[HTML]{000000}{\tiny 24}}} & \colorbox[HTML]{FFEBDA}{\makebox[0.58cm][c]{\textcolor[HTML]{000000}{\tiny 18.2}}} & \makebox[0.58cm][c]{\tiny ---} & \makebox[0.52cm][c]{\tiny ---} & \makebox[0.46cm][c]{\tiny ---} \\
 \multirow{1}{*}{\textit{Stereo}} & FS~\cite{wen2025foundationstereo} {\tiny (fitted)} & 1 & 2 & \colorbox[HTML]{66A574}{\makebox[0.46cm][c]{\textcolor[HTML]{000000}{\tiny 39}}} & \colorbox[HTML]{8B3200}{\makebox[0.58cm][c]{\textcolor[HTML]{FFFFFF}{\tiny 1.6}}} & \makebox[0.58cm][c]{\tiny ---} & \makebox[0.52cm][c]{\tiny ---} & \makebox[0.46cm][c]{\tiny ---} \\
\multirow{1}{*}{\textit{Stereo}} & FS~\cite{wen2025foundationstereo} {\tiny (anchored)} & 1 & 2 & \colorbox[HTML]{66A574}{\makebox[0.46cm][c]{\textcolor[HTML]{000000}{\tiny 36}}} & \colorbox[HTML]{8B3200}{\makebox[0.58cm][c]{\textcolor[HTML]{FFFFFF}{\tiny 1.6}}} & \makebox[0.58cm][c]{\tiny ---} & \makebox[0.52cm][c]{\tiny ---} & \makebox[0.46cm][c]{\tiny ---} \\
\specialrule{0.9pt}{1.5pt}{1.5pt}
\multirow{2}{*}{\textit{Multi-stereo}} & FS~\cite{wen2025foundationstereo} + SE(3) reg. & 2 & 4 & \colorbox[HTML]{478F59}{\makebox[0.46cm][c]{\textcolor[HTML]{FFFFFF}{\tiny 44}}} & \colorbox[HTML]{8B3200}{\makebox[0.58cm][c]{\textcolor[HTML]{FFFFFF}{\tiny 1.6}}} & \colorbox[HTML]{D89B7D}{\makebox[0.58cm][c]{\textcolor[HTML]{000000}{\tiny 6.3}}} & \colorbox[HTML]{C2DDFF}{\makebox[0.52cm][c]{\textcolor[HTML]{000000}{\tiny 12.1}}} & \colorbox[HTML]{A4C5F1}{\makebox[0.46cm][c]{\textcolor[HTML]{000000}{\tiny 2.7}}} \\
 & FS~\cite{wen2025foundationstereo} + Sim(3) reg. & 2 & 4 & \colorbox[HTML]{237B3F}{\makebox[0.46cm][c]{\textcolor[HTML]{FFFFFF}{\tiny 50}}} & \colorbox[HTML]{8B3200}{\makebox[0.58cm][c]{\textcolor[HTML]{FFFFFF}{\tiny 1.6}}} & {\definecolor{ldbg}{HTML}{B26740}%
\tikz[baseline=(ldc.base)]{\node[inner sep=1.2pt,outer sep=0pt,line width=0.6pt,draw=black,dash pattern=on 1.1pt off 0.9pt,fill=ldbg] (ldc) {\makebox[0.58cm][c]{\textcolor[HTML]{FFFFFF}{\tiny 2.8}}};}} & \colorbox[HTML]{6A96CF}{\makebox[0.52cm][c]{\textcolor[HTML]{000000}{\tiny 4.6}}} & \colorbox[HTML]{87AEE0}{\makebox[0.46cm][c]{\textcolor[HTML]{000000}{\tiny 1.7}}} \\
\multirow{1}{*}{\textit{Multi-view}} & DA3~\cite{lin2025depth} + GGPT~\cite{chen2026ggpt} & 2 & 4 & \colorbox[HTML]{237B3F}{\makebox[0.46cm][c]{\textcolor[HTML]{FFFFFF}{\tiny 48}}} & \colorbox[HTML]{B26740}{\makebox[0.58cm][c]{\textcolor[HTML]{FFFFFF}{\tiny 3.4}}} & \colorbox[HTML]{D89B7D}{\makebox[0.58cm][c]{\textcolor[HTML]{000000}{\tiny 5.3}}} & \colorbox[HTML]{003A8A}{\makebox[0.52cm][c]{\textcolor[HTML]{FFFFFF}{\tiny 0.9}}} & \colorbox[HTML]{003A8A}{\makebox[0.46cm][c]{\textcolor[HTML]{FFFFFF}{\tiny 0.3}}} \\
\specialrule{0.9pt}{1.5pt}{1.5pt}
\multirow{2}{*}{\textit{Multi-stereo}} & FS~\cite{wen2025foundationstereo} + SE(3) reg. & 3 & 6 & \colorbox[HTML]{237B3F}{\makebox[0.46cm][c]{\textcolor[HTML]{FFFFFF}{\tiny 52}}} & \colorbox[HTML]{8B3200}{\makebox[0.58cm][c]{\textcolor[HTML]{FFFFFF}{\tiny 1.6}}} & \colorbox[HTML]{D89B7D}{\makebox[0.58cm][c]{\textcolor[HTML]{000000}{\tiny 6.3}}} & \colorbox[HTML]{A4C5F1}{\makebox[0.52cm][c]{\textcolor[HTML]{000000}{\tiny 10.6}}} & \colorbox[HTML]{87AEE0}{\makebox[0.46cm][c]{\textcolor[HTML]{000000}{\tiny 2.1}}} \\
 & FS~\cite{wen2025foundationstereo} + Sim(3) reg. & 3 & 6 & \colorbox[HTML]{006624}{\makebox[0.46cm][c]{\textcolor[HTML]{FFFFFF}{\tiny 58}}} & \colorbox[HTML]{8B3200}{\makebox[0.58cm][c]{\textcolor[HTML]{FFFFFF}{\tiny 1.6}}} & \colorbox[HTML]{B26740}{\makebox[0.58cm][c]{\textcolor[HTML]{FFFFFF}{\tiny 2.8}}} & \colorbox[HTML]{6A96CF}{\makebox[0.52cm][c]{\textcolor[HTML]{000000}{\tiny 3.7}}} & \colorbox[HTML]{6A96CF}{\makebox[0.46cm][c]{\textcolor[HTML]{000000}{\tiny 1.4}}} \\
\multirow{1}{*}{\textit{Multi-view}} & DA3~\cite{lin2025depth} + GGPT~\cite{chen2026ggpt} & 3 & 6 & {\definecolor{ldbg}{HTML}{005202}%
\tikz[baseline=(ldc.base)]{\node[inner sep=1.2pt,outer sep=0pt,line width=0.6pt,draw=black,dash pattern=on 1.1pt off 0.9pt,fill=ldbg] (ldc) {\makebox[0.46cm][c]{\textcolor[HTML]{FFFFFF}{\tiny 67}}};}} & {\definecolor{ldbg}{HTML}{781100}%
\tikz[baseline=(ldc.base)]{\node[inner sep=1.2pt,outer sep=0pt,line width=0.6pt,draw=black,dash pattern=on 1.1pt off 0.9pt,fill=ldbg] (ldc) {\makebox[0.58cm][c]{\textcolor[HTML]{FFFFFF}{\tiny 1.2}}};}} & \colorbox[HTML]{C5815F}{\makebox[0.58cm][c]{\textcolor[HTML]{000000}{\tiny 3.8}}} & {\definecolor{ldbg}{HTML}{003A8A}%
\tikz[baseline=(ldc.base)]{\node[inner sep=1.2pt,outer sep=0pt,line width=0.6pt,draw=black,dash pattern=on 1.1pt off 0.9pt,fill=ldbg] (ldc) {\makebox[0.52cm][c]{\textcolor[HTML]{FFFFFF}{\tiny 0.6}}};}} & {\definecolor{ldbg}{HTML}{003A8A}%
\tikz[baseline=(ldc.base)]{\node[inner sep=1.2pt,outer sep=0pt,line width=0.6pt,draw=black,dash pattern=on 1.1pt off 0.9pt,fill=ldbg] (ldc) {\makebox[0.46cm][c]{\textcolor[HTML]{FFFFFF}{\tiny 0.2}}};}} \\
\end{tabular}
\end{table}

\begin{table*}[t!]
\centering\scriptsize
\setlength{\fboxsep}{1.5pt}
\setlength{\tabcolsep}{1.45pt}\renewcommand{\arraystretch}{1.25}
\caption{Comparison of multi-view methods across viewpoint count. Mean error (ME) in the reference view E1L measures geometric accuracy in a common image region, while E2L additionally reflects cross-camera pose and view overlap. Geometry-refined methods benefit most from the third viewpoint and achieve the highest coverage (Cov) and lowest E1L error at six synchronized views. Darker shading indicates better performance; outlined cells denote the best value in each column.}
\label{tab:results_fused_split}
\begin{tabular}{l@{\hspace{3pt}} @{}c@{\hspace{3.5pt}} cccccccc !{\vrule width 1pt} cccc !{\vrule width 1pt} cccccccc}
& & \multicolumn{8}{c}{\textbf{Mean Error}\,$\downarrow$} & \multicolumn{4}{c}{\textbf{Coverage of GT}\,$\uparrow$} & \multicolumn{8}{c}{\textbf{Pose Error}\,$\downarrow$} \\
\cmidrule(lr){3-10}\cmidrule(lr){11-14}\cmidrule(lr){15-22}
& & \multicolumn{4}{c}{\textbf{Per-view} {\tiny (mm) @\textbf{E1L}}} & \multicolumn{4}{c}{\textbf{Per-view} {\tiny (mm) @\textbf{E2L}}} & \multicolumn{4}{c}{\textbf{All views} {\tiny (\%) @\textbf{1mm}}} & \multicolumn{4}{c}{\textbf{Translation} {\tiny (mm) @\textbf{left pairs}}} & \multicolumn{4}{c}{\textbf{Rotation} {\tiny (deg) @\textbf{left pairs}}} \\
\cmidrule(lr){3-6}\cmidrule(lr){7-10}\cmidrule(lr){11-14}\cmidrule(lr){15-18}\cmidrule(lr){19-22}
\noalign{\makeatletter\global\setbox\@arstrutbox\hbox{\vrule\@height0.62\ht\strutbox\@depth0.62\dp\strutbox\@width\z@}\makeatother}
\multicolumn{2}{r@{\hspace{3.5pt}}}{\scriptsize\bfseries Viewpoints $\rightarrow$} & \multicolumn{2}{c}{\scriptsize Two} & \multicolumn{2}{c}{\scriptsize Three} & \multicolumn{2}{c}{\scriptsize Two} & \multicolumn{2}{c}{\scriptsize Three} & \multicolumn{2}{c}{\scriptsize Two} & \multicolumn{2}{c}{\scriptsize Three} & \multicolumn{2}{c}{\scriptsize Two} & \multicolumn{2}{c}{\scriptsize Three} & \multicolumn{2}{c}{\scriptsize Two} & \multicolumn{2}{c}{\scriptsize Three} \\[-1pt]
\cmidrule[0.3pt](lr){3-4}\cmidrule[0.3pt](lr){5-6}\cmidrule[0.3pt](lr){7-8}\cmidrule[0.3pt](lr){9-10}\cmidrule[0.3pt](lr){11-12}\cmidrule[0.3pt](lr){13-14}\cmidrule[0.3pt](lr){15-16}\cmidrule[0.3pt](lr){17-18}\cmidrule[0.3pt](lr){19-20}\cmidrule[0.3pt](lr){21-22}
\multicolumn{2}{r@{\hspace{3.5pt}}}{\scriptsize\bfseries Synch.\ Views $\rightarrow$} & {\scriptsize 2} & {\scriptsize 4} & {\scriptsize 3} & {\scriptsize 6} & {\scriptsize 2} & {\scriptsize 4} & {\scriptsize 3} & \multicolumn{1}{c}{{\scriptsize 6}} & {\scriptsize 2} & {\scriptsize 4} & {\scriptsize 3} & \multicolumn{1}{c}{{\scriptsize 6}} & {\scriptsize 2} & {\scriptsize 4} & {\scriptsize 3} & {\scriptsize 6} & {\scriptsize 2} & {\scriptsize 4} & {\scriptsize 3} & {\scriptsize 6} \\[-0.5pt]
\noalign{\makeatletter\global\setbox\@arstrutbox\hbox{\vrule\@height1.25\ht\strutbox\@depth1.25\dp\strutbox\@width\z@}\makeatother}
\midrule
VGGT~\cite{wang2025vggt} & \multirow{4}{*}{\rotatebox[origin=c]{90}{\textit{(FF)}}} & \colorbox[HTML]{FCD0BB}{\makebox[0.59cm][c]{\textcolor[HTML]{000000}{\tiny 14.6}}} & \colorbox[HTML]{FCD0BB}{\makebox[0.59cm][c]{\textcolor[HTML]{000000}{\tiny 14.6}}} & \colorbox[HTML]{EAB59B}{\makebox[0.59cm][c]{\textcolor[HTML]{000000}{\tiny 8.4}}} & \colorbox[HTML]{EAB59B}{\makebox[0.59cm][c]{\textcolor[HTML]{000000}{\tiny 8.9}}} & \colorbox[HTML]{FFEBDA}{\makebox[0.59cm][c]{\textcolor[HTML]{000000}{\tiny 18.9}}} & \colorbox[HTML]{FFEBDA}{\makebox[0.59cm][c]{\textcolor[HTML]{000000}{\tiny 20.8}}} & \colorbox[HTML]{EAB59B}{\makebox[0.59cm][c]{\textcolor[HTML]{000000}{\tiny 10.5}}} & \colorbox[HTML]{FCD0BB}{\makebox[0.59cm][c]{\textcolor[HTML]{000000}{\tiny 11.1}}} & \colorbox[HTML]{DFFDE4}{\makebox[0.44cm][c]{\textcolor[HTML]{000000}{\tiny 8}}} & \colorbox[HTML]{DFFDE4}{\makebox[0.44cm][c]{\textcolor[HTML]{000000}{\tiny 9}}} & \colorbox[HTML]{A2D0AA}{\makebox[0.44cm][c]{\textcolor[HTML]{000000}{\tiny 25}}} & \colorbox[HTML]{84BA8F}{\makebox[0.44cm][c]{\textcolor[HTML]{000000}{\tiny 28}}} & \colorbox[HTML]{E0F6FF}{\makebox[0.54cm][c]{\textcolor[HTML]{000000}{\tiny 23.9}}} & \colorbox[HTML]{E0F6FF}{\makebox[0.54cm][c]{\textcolor[HTML]{000000}{\tiny 26.6}}} & \colorbox[HTML]{C2DDFF}{\makebox[0.54cm][c]{\textcolor[HTML]{000000}{\tiny 11.7}}} & \colorbox[HTML]{C2DDFF}{\makebox[0.54cm][c]{\textcolor[HTML]{000000}{\tiny 12.3}}} & \colorbox[HTML]{E0F6FF}{\makebox[0.44cm][c]{\textcolor[HTML]{000000}{\tiny 5.7}}} & \colorbox[HTML]{E0F6FF}{\makebox[0.44cm][c]{\textcolor[HTML]{000000}{\tiny 6.3}}} & \colorbox[HTML]{A4C5F1}{\makebox[0.44cm][c]{\textcolor[HTML]{000000}{\tiny 2.8}}} & \colorbox[HTML]{A4C5F1}{\makebox[0.44cm][c]{\textcolor[HTML]{000000}{\tiny 2.9}}} \\
DA3~\cite{lin2025depth} &  & \colorbox[HTML]{EAB59B}{\makebox[0.59cm][c]{\textcolor[HTML]{000000}{\tiny 8.5}}} & \colorbox[HTML]{EAB59B}{\makebox[0.59cm][c]{\textcolor[HTML]{000000}{\tiny 8.5}}} & \colorbox[HTML]{C5815F}{\makebox[0.59cm][c]{\textcolor[HTML]{000000}{\tiny 4.6}}} & \colorbox[HTML]{C5815F}{\makebox[0.59cm][c]{\textcolor[HTML]{000000}{\tiny 4.8}}} & \colorbox[HTML]{FCD0BB}{\makebox[0.59cm][c]{\textcolor[HTML]{000000}{\tiny 11.7}}} & \colorbox[HTML]{FCD0BB}{\makebox[0.59cm][c]{\textcolor[HTML]{000000}{\tiny 12.2}}} & \colorbox[HTML]{D89B7D}{\makebox[0.59cm][c]{\textcolor[HTML]{000000}{\tiny 7.0}}} & \colorbox[HTML]{D89B7D}{\makebox[0.59cm][c]{\textcolor[HTML]{000000}{\tiny 7.6}}} & \colorbox[HTML]{C0E6C7}{\makebox[0.44cm][c]{\textcolor[HTML]{000000}{\tiny 17}}} & \colorbox[HTML]{A2D0AA}{\makebox[0.44cm][c]{\textcolor[HTML]{000000}{\tiny 21}}} & \colorbox[HTML]{66A574}{\makebox[0.44cm][c]{\textcolor[HTML]{000000}{\tiny 37}}} & \colorbox[HTML]{478F59}{\makebox[0.44cm][c]{\textcolor[HTML]{FFFFFF}{\tiny 42}}} & \colorbox[HTML]{E0F6FF}{\makebox[0.54cm][c]{\textcolor[HTML]{000000}{\tiny 20.2}}} & \colorbox[HTML]{E0F6FF}{\makebox[0.54cm][c]{\textcolor[HTML]{000000}{\tiny 19.8}}} & \colorbox[HTML]{A4C5F1}{\makebox[0.54cm][c]{\textcolor[HTML]{000000}{\tiny 9.9}}} & \colorbox[HTML]{A4C5F1}{\makebox[0.54cm][c]{\textcolor[HTML]{000000}{\tiny 9.4}}} & \colorbox[HTML]{A4C5F1}{\makebox[0.44cm][c]{\textcolor[HTML]{000000}{\tiny 3.5}}} & \colorbox[HTML]{A4C5F1}{\makebox[0.44cm][c]{\textcolor[HTML]{000000}{\tiny 3.1}}} & \colorbox[HTML]{6A96CF}{\makebox[0.44cm][c]{\textcolor[HTML]{000000}{\tiny 1.5}}} & \colorbox[HTML]{87AEE0}{\makebox[0.44cm][c]{\textcolor[HTML]{000000}{\tiny 1.7}}} \\
Pi3~\cite{wang2025pi} &  & \colorbox[HTML]{C5815F}{\makebox[0.59cm][c]{\textcolor[HTML]{000000}{\tiny 4.9}}} & \colorbox[HTML]{C5815F}{\makebox[0.59cm][c]{\textcolor[HTML]{000000}{\tiny 4.6}}} & \colorbox[HTML]{B26740}{\makebox[0.59cm][c]{\textcolor[HTML]{FFFFFF}{\tiny 3.7}}} & \colorbox[HTML]{B26740}{\makebox[0.59cm][c]{\textcolor[HTML]{FFFFFF}{\tiny 3.4}}} & \colorbox[HTML]{EAB59B}{\makebox[0.59cm][c]{\textcolor[HTML]{000000}{\tiny 8.7}}} & \colorbox[HTML]{EAB59B}{\makebox[0.59cm][c]{\textcolor[HTML]{000000}{\tiny 8.1}}} & \colorbox[HTML]{D89B7D}{\makebox[0.59cm][c]{\textcolor[HTML]{000000}{\tiny 6.9}}} & \colorbox[HTML]{D89B7D}{\makebox[0.59cm][c]{\textcolor[HTML]{000000}{\tiny 6.0}}} & \colorbox[HTML]{A2D0AA}{\makebox[0.44cm][c]{\textcolor[HTML]{000000}{\tiny 22}}} & \colorbox[HTML]{A2D0AA}{\makebox[0.44cm][c]{\textcolor[HTML]{000000}{\tiny 27}}} & \colorbox[HTML]{66A574}{\makebox[0.44cm][c]{\textcolor[HTML]{000000}{\tiny 38}}} & \colorbox[HTML]{478F59}{\makebox[0.44cm][c]{\textcolor[HTML]{FFFFFF}{\tiny 44}}} & \colorbox[HTML]{E0F6FF}{\makebox[0.54cm][c]{\textcolor[HTML]{000000}{\tiny 22.5}}} & \colorbox[HTML]{E0F6FF}{\makebox[0.54cm][c]{\textcolor[HTML]{000000}{\tiny 21.6}}} & \colorbox[HTML]{C2DDFF}{\makebox[0.54cm][c]{\textcolor[HTML]{000000}{\tiny 11.5}}} & \colorbox[HTML]{A4C5F1}{\makebox[0.54cm][c]{\textcolor[HTML]{000000}{\tiny 11.3}}} & \colorbox[HTML]{C2DDFF}{\makebox[0.44cm][c]{\textcolor[HTML]{000000}{\tiny 4.0}}} & \colorbox[HTML]{C2DDFF}{\makebox[0.44cm][c]{\textcolor[HTML]{000000}{\tiny 3.9}}} & \colorbox[HTML]{87AEE0}{\makebox[0.44cm][c]{\textcolor[HTML]{000000}{\tiny 2.3}}} & \colorbox[HTML]{87AEE0}{\makebox[0.44cm][c]{\textcolor[HTML]{000000}{\tiny 2.1}}} \\
Pi3X~\cite{wang2025pi3x} &  & \colorbox[HTML]{C5815F}{\makebox[0.59cm][c]{\textcolor[HTML]{000000}{\tiny 4.4}}} & \colorbox[HTML]{B26740}{\makebox[0.59cm][c]{\textcolor[HTML]{FFFFFF}{\tiny 3.8}}} & \colorbox[HTML]{B26740}{\makebox[0.59cm][c]{\textcolor[HTML]{FFFFFF}{\tiny 3.0}}} & \colorbox[HTML]{9F4D1E}{\makebox[0.59cm][c]{\textcolor[HTML]{FFFFFF}{\tiny 2.6}}} & \colorbox[HTML]{FCD0BB}{\makebox[0.59cm][c]{\textcolor[HTML]{000000}{\tiny 11.8}}} & \colorbox[HTML]{FCD0BB}{\makebox[0.59cm][c]{\textcolor[HTML]{000000}{\tiny 11.3}}} & \colorbox[HTML]{EAB59B}{\makebox[0.59cm][c]{\textcolor[HTML]{000000}{\tiny 9.1}}} & \colorbox[HTML]{EAB59B}{\makebox[0.59cm][c]{\textcolor[HTML]{000000}{\tiny 9.1}}} & \colorbox[HTML]{A2D0AA}{\makebox[0.44cm][c]{\textcolor[HTML]{000000}{\tiny 24}}} & \colorbox[HTML]{84BA8F}{\makebox[0.44cm][c]{\textcolor[HTML]{000000}{\tiny 29}}} & \colorbox[HTML]{66A574}{\makebox[0.44cm][c]{\textcolor[HTML]{000000}{\tiny 38}}} & \colorbox[HTML]{478F59}{\makebox[0.44cm][c]{\textcolor[HTML]{FFFFFF}{\tiny 43}}} & \colorbox[HTML]{C2DDFF}{\makebox[0.54cm][c]{\textcolor[HTML]{000000}{\tiny 16.7}}} & \colorbox[HTML]{C2DDFF}{\makebox[0.54cm][c]{\textcolor[HTML]{000000}{\tiny 17.2}}} & \colorbox[HTML]{A4C5F1}{\makebox[0.54cm][c]{\textcolor[HTML]{000000}{\tiny 8.1}}} & \colorbox[HTML]{A4C5F1}{\makebox[0.54cm][c]{\textcolor[HTML]{000000}{\tiny 8.4}}} & \colorbox[HTML]{E0F6FF}{\makebox[0.44cm][c]{\textcolor[HTML]{000000}{\tiny 7.4}}} & \colorbox[HTML]{E0F6FF}{\makebox[0.44cm][c]{\textcolor[HTML]{000000}{\tiny 7.7}}} & \colorbox[HTML]{A4C5F1}{\makebox[0.44cm][c]{\textcolor[HTML]{000000}{\tiny 3.3}}} & \colorbox[HTML]{A4C5F1}{\makebox[0.44cm][c]{\textcolor[HTML]{000000}{\tiny 3.3}}} \\
\specialrule{0.9pt}{1.5pt}{1.5pt}
MASt3R-SfM~\cite{duisterhof2025mast3r} & \multirow{4}{*}{\rotatebox[origin=c]{90}{\textit{(FF+\emph{SfM})}}} & \colorbox[HTML]{EAB59B}{\makebox[0.59cm][c]{\textcolor[HTML]{000000}{\tiny 7.9}}} & \colorbox[HTML]{B26740}{\makebox[0.59cm][c]{\textcolor[HTML]{FFFFFF}{\tiny 3.8}}} & \colorbox[HTML]{B26740}{\makebox[0.59cm][c]{\textcolor[HTML]{FFFFFF}{\tiny 3.6}}} & \colorbox[HTML]{9F4D1E}{\makebox[0.59cm][c]{\textcolor[HTML]{FFFFFF}{\tiny 2.7}}} & \colorbox[HTML]{EAB59B}{\makebox[0.59cm][c]{\textcolor[HTML]{000000}{\tiny 10.1}}} & {\definecolor{ldbg}{HTML}{C5815F}%
\tikz[baseline=(ldc.base)]{\node[inner sep=1.2pt,outer sep=0pt,line width=0.6pt,draw=black,dash pattern=on 1.1pt off 0.9pt,fill=ldbg] (ldc) {\makebox[0.59cm][c]{\textcolor[HTML]{000000}{\tiny 5.1}}};}} & \colorbox[HTML]{EAB59B}{\makebox[0.59cm][c]{\textcolor[HTML]{000000}{\tiny 9.9}}} & \colorbox[HTML]{C5815F}{\makebox[0.59cm][c]{\textcolor[HTML]{000000}{\tiny 4.8}}} & \colorbox[HTML]{84BA8F}{\makebox[0.44cm][c]{\textcolor[HTML]{000000}{\tiny 31}}} & {\definecolor{ldbg}{HTML}{237B3F}%
\tikz[baseline=(ldc.base)]{\node[inner sep=1.2pt,outer sep=0pt,line width=0.6pt,draw=black,dash pattern=on 1.1pt off 0.9pt,fill=ldbg] (ldc) {\makebox[0.44cm][c]{\textcolor[HTML]{FFFFFF}{\tiny 49}}};}} & \colorbox[HTML]{237B3F}{\makebox[0.44cm][c]{\textcolor[HTML]{FFFFFF}{\tiny 48}}} & \colorbox[HTML]{006624}{\makebox[0.44cm][c]{\textcolor[HTML]{FFFFFF}{\tiny 59}}} & \colorbox[HTML]{87AEE0}{\makebox[0.54cm][c]{\textcolor[HTML]{000000}{\tiny 5.3}}} & \colorbox[HTML]{87AEE0}{\makebox[0.54cm][c]{\textcolor[HTML]{000000}{\tiny 5.2}}} & \colorbox[HTML]{4E7FBE}{\makebox[0.54cm][c]{\textcolor[HTML]{FFFFFF}{\tiny 3.1}}} & \colorbox[HTML]{4E7FBE}{\makebox[0.54cm][c]{\textcolor[HTML]{FFFFFF}{\tiny 2.9}}} & \colorbox[HTML]{A4C5F1}{\makebox[0.44cm][c]{\textcolor[HTML]{000000}{\tiny 3.4}}} & \colorbox[HTML]{A4C5F1}{\makebox[0.44cm][c]{\textcolor[HTML]{000000}{\tiny 2.7}}} & \colorbox[HTML]{87AEE0}{\makebox[0.44cm][c]{\textcolor[HTML]{000000}{\tiny 1.7}}} & \colorbox[HTML]{6A96CF}{\makebox[0.44cm][c]{\textcolor[HTML]{000000}{\tiny 1.6}}} \\
VGGT~\cite{wang2025vggt} + GGPT~\cite{chen2026ggpt} &  & \colorbox[HTML]{C5815F}{\makebox[0.59cm][c]{\textcolor[HTML]{000000}{\tiny 4.3}}} & \colorbox[HTML]{B26740}{\makebox[0.59cm][c]{\textcolor[HTML]{FFFFFF}{\tiny 4.0}}} & \colorbox[HTML]{8B3200}{\makebox[0.59cm][c]{\textcolor[HTML]{FFFFFF}{\tiny 1.6}}} & \colorbox[HTML]{781100}{\makebox[0.59cm][c]{\textcolor[HTML]{FFFFFF}{\tiny 1.4}}} & \colorbox[HTML]{D89B7D}{\makebox[0.59cm][c]{\textcolor[HTML]{000000}{\tiny 6.4}}} & \colorbox[HTML]{D89B7D}{\makebox[0.59cm][c]{\textcolor[HTML]{000000}{\tiny 7.1}}} & \colorbox[HTML]{C5815F}{\makebox[0.59cm][c]{\textcolor[HTML]{000000}{\tiny 4.3}}} & {\definecolor{ldbg}{HTML}{B26740}%
\tikz[baseline=(ldc.base)]{\node[inner sep=1.2pt,outer sep=0pt,line width=0.6pt,draw=black,dash pattern=on 1.1pt off 0.9pt,fill=ldbg] (ldc) {\makebox[0.59cm][c]{\textcolor[HTML]{FFFFFF}{\tiny 3.7}}};}} & {\definecolor{ldbg}{HTML}{478F59}%
\tikz[baseline=(ldc.base)]{\node[inner sep=1.2pt,outer sep=0pt,line width=0.6pt,draw=black,dash pattern=on 1.1pt off 0.9pt,fill=ldbg] (ldc) {\makebox[0.44cm][c]{\textcolor[HTML]{FFFFFF}{\tiny 41}}};}} & \colorbox[HTML]{478F59}{\makebox[0.44cm][c]{\textcolor[HTML]{FFFFFF}{\tiny 46}}} & \colorbox[HTML]{005202}{\makebox[0.44cm][c]{\textcolor[HTML]{FFFFFF}{\tiny 63}}} & \colorbox[HTML]{005202}{\makebox[0.44cm][c]{\textcolor[HTML]{FFFFFF}{\tiny 67}}} & \colorbox[HTML]{0C519B}{\makebox[0.54cm][c]{\textcolor[HTML]{FFFFFF}{\tiny 1.0}}} & \colorbox[HTML]{4E7FBE}{\makebox[0.54cm][c]{\textcolor[HTML]{FFFFFF}{\tiny 3.1}}} & {\definecolor{ldbg}{HTML}{003A8A}%
\tikz[baseline=(ldc.base)]{\node[inner sep=1.2pt,outer sep=0pt,line width=0.6pt,draw=black,dash pattern=on 1.1pt off 0.9pt,fill=ldbg] (ldc) {\makebox[0.54cm][c]{\textcolor[HTML]{FFFFFF}{\tiny 0.6}}};}} & \colorbox[HTML]{003A8A}{\makebox[0.54cm][c]{\textcolor[HTML]{FFFFFF}{\tiny 0.6}}} & {\definecolor{ldbg}{HTML}{0C519B}%
\tikz[baseline=(ldc.base)]{\node[inner sep=1.2pt,outer sep=0pt,line width=0.6pt,draw=black,dash pattern=on 1.1pt off 0.9pt,fill=ldbg] (ldc) {\makebox[0.44cm][c]{\textcolor[HTML]{FFFFFF}{\tiny 0.3}}};}} & \colorbox[HTML]{6A96CF}{\makebox[0.44cm][c]{\textcolor[HTML]{000000}{\tiny 1.3}}} & \colorbox[HTML]{003A8A}{\makebox[0.44cm][c]{\textcolor[HTML]{FFFFFF}{\tiny 0.3}}} & \colorbox[HTML]{003A8A}{\makebox[0.44cm][c]{\textcolor[HTML]{FFFFFF}{\tiny 0.2}}} \\
Pi3X~\cite{wang2025pi3x} + GGPT~\cite{chen2026ggpt} &  & \colorbox[HTML]{C5815F}{\makebox[0.59cm][c]{\textcolor[HTML]{000000}{\tiny 4.6}}} & {\definecolor{ldbg}{HTML}{9F4D1E}%
\tikz[baseline=(ldc.base)]{\node[inner sep=1.2pt,outer sep=0pt,line width=0.6pt,draw=black,dash pattern=on 1.1pt off 0.9pt,fill=ldbg] (ldc) {\makebox[0.59cm][c]{\textcolor[HTML]{FFFFFF}{\tiny 2.7}}};}} & \colorbox[HTML]{781100}{\makebox[0.59cm][c]{\textcolor[HTML]{FFFFFF}{\tiny 1.3}}} & {\definecolor{ldbg}{HTML}{781100}%
\tikz[baseline=(ldc.base)]{\node[inner sep=1.2pt,outer sep=0pt,line width=0.6pt,draw=black,dash pattern=on 1.1pt off 0.9pt,fill=ldbg] (ldc) {\makebox[0.59cm][c]{\textcolor[HTML]{FFFFFF}{\tiny 1.1}}};}} & \colorbox[HTML]{EAB59B}{\makebox[0.59cm][c]{\textcolor[HTML]{000000}{\tiny 7.9}}} & \colorbox[HTML]{EAB59B}{\makebox[0.59cm][c]{\textcolor[HTML]{000000}{\tiny 8.2}}} & \colorbox[HTML]{D89B7D}{\makebox[0.59cm][c]{\textcolor[HTML]{000000}{\tiny 6.0}}} & \colorbox[HTML]{C5815F}{\makebox[0.59cm][c]{\textcolor[HTML]{000000}{\tiny 5.0}}} & \colorbox[HTML]{66A574}{\makebox[0.44cm][c]{\textcolor[HTML]{000000}{\tiny 40}}} & \colorbox[HTML]{478F59}{\makebox[0.44cm][c]{\textcolor[HTML]{FFFFFF}{\tiny 46}}} & \colorbox[HTML]{005202}{\makebox[0.44cm][c]{\textcolor[HTML]{FFFFFF}{\tiny 62}}} & \colorbox[HTML]{005202}{\makebox[0.44cm][c]{\textcolor[HTML]{FFFFFF}{\tiny 66}}} & \colorbox[HTML]{6A96CF}{\makebox[0.54cm][c]{\textcolor[HTML]{000000}{\tiny 4.6}}} & \colorbox[HTML]{3068AD}{\makebox[0.54cm][c]{\textcolor[HTML]{FFFFFF}{\tiny 1.6}}} & \colorbox[HTML]{3068AD}{\makebox[0.54cm][c]{\textcolor[HTML]{FFFFFF}{\tiny 1.4}}} & {\definecolor{ldbg}{HTML}{003A8A}%
\tikz[baseline=(ldc.base)]{\node[inner sep=1.2pt,outer sep=0pt,line width=0.6pt,draw=black,dash pattern=on 1.1pt off 0.9pt,fill=ldbg] (ldc) {\makebox[0.54cm][c]{\textcolor[HTML]{FFFFFF}{\tiny 0.6}}};}} & \colorbox[HTML]{87AEE0}{\makebox[0.44cm][c]{\textcolor[HTML]{000000}{\tiny 2.3}}} & \colorbox[HTML]{4E7FBE}{\makebox[0.44cm][c]{\textcolor[HTML]{FFFFFF}{\tiny 0.8}}} & \colorbox[HTML]{3068AD}{\makebox[0.44cm][c]{\textcolor[HTML]{FFFFFF}{\tiny 0.6}}} & {\definecolor{ldbg}{HTML}{003A8A}%
\tikz[baseline=(ldc.base)]{\node[inner sep=1.2pt,outer sep=0pt,line width=0.6pt,draw=black,dash pattern=on 1.1pt off 0.9pt,fill=ldbg] (ldc) {\makebox[0.44cm][c]{\textcolor[HTML]{FFFFFF}{\tiny 0.2}}};}} \\
DA3~\cite{lin2025depth} + GGPT~\cite{chen2026ggpt} &  & {\definecolor{ldbg}{HTML}{C5815F}%
\tikz[baseline=(ldc.base)]{\node[inner sep=1.2pt,outer sep=0pt,line width=0.6pt,draw=black,dash pattern=on 1.1pt off 0.9pt,fill=ldbg] (ldc) {\makebox[0.59cm][c]{\textcolor[HTML]{000000}{\tiny 4.3}}};}} & \colorbox[HTML]{B26740}{\makebox[0.59cm][c]{\textcolor[HTML]{FFFFFF}{\tiny 3.4}}} & {\definecolor{ldbg}{HTML}{781100}%
\tikz[baseline=(ldc.base)]{\node[inner sep=1.2pt,outer sep=0pt,line width=0.6pt,draw=black,dash pattern=on 1.1pt off 0.9pt,fill=ldbg] (ldc) {\makebox[0.59cm][c]{\textcolor[HTML]{FFFFFF}{\tiny 1.3}}};}} & \colorbox[HTML]{781100}{\makebox[0.59cm][c]{\textcolor[HTML]{FFFFFF}{\tiny 1.2}}} & {\definecolor{ldbg}{HTML}{D89B7D}%
\tikz[baseline=(ldc.base)]{\node[inner sep=1.2pt,outer sep=0pt,line width=0.6pt,draw=black,dash pattern=on 1.1pt off 0.9pt,fill=ldbg] (ldc) {\makebox[0.59cm][c]{\textcolor[HTML]{000000}{\tiny 5.6}}};}} & \colorbox[HTML]{C5815F}{\makebox[0.59cm][c]{\textcolor[HTML]{000000}{\tiny 5.3}}} & {\definecolor{ldbg}{HTML}{C5815F}%
\tikz[baseline=(ldc.base)]{\node[inner sep=1.2pt,outer sep=0pt,line width=0.6pt,draw=black,dash pattern=on 1.1pt off 0.9pt,fill=ldbg] (ldc) {\makebox[0.59cm][c]{\textcolor[HTML]{000000}{\tiny 4.2}}};}} & \colorbox[HTML]{B26740}{\makebox[0.59cm][c]{\textcolor[HTML]{FFFFFF}{\tiny 3.8}}} & \colorbox[HTML]{478F59}{\makebox[0.44cm][c]{\textcolor[HTML]{FFFFFF}{\tiny 41}}} & \colorbox[HTML]{237B3F}{\makebox[0.44cm][c]{\textcolor[HTML]{FFFFFF}{\tiny 48}}} & {\definecolor{ldbg}{HTML}{005202}%
\tikz[baseline=(ldc.base)]{\node[inner sep=1.2pt,outer sep=0pt,line width=0.6pt,draw=black,dash pattern=on 1.1pt off 0.9pt,fill=ldbg] (ldc) {\makebox[0.44cm][c]{\textcolor[HTML]{FFFFFF}{\tiny 63}}};}} & {\definecolor{ldbg}{HTML}{005202}%
\tikz[baseline=(ldc.base)]{\node[inner sep=1.2pt,outer sep=0pt,line width=0.6pt,draw=black,dash pattern=on 1.1pt off 0.9pt,fill=ldbg] (ldc) {\makebox[0.44cm][c]{\textcolor[HTML]{FFFFFF}{\tiny 67}}};}} & {\definecolor{ldbg}{HTML}{0C519B}%
\tikz[baseline=(ldc.base)]{\node[inner sep=1.2pt,outer sep=0pt,line width=0.6pt,draw=black,dash pattern=on 1.1pt off 0.9pt,fill=ldbg] (ldc) {\makebox[0.54cm][c]{\textcolor[HTML]{FFFFFF}{\tiny 1.0}}};}} & {\definecolor{ldbg}{HTML}{003A8A}%
\tikz[baseline=(ldc.base)]{\node[inner sep=1.2pt,outer sep=0pt,line width=0.6pt,draw=black,dash pattern=on 1.1pt off 0.9pt,fill=ldbg] (ldc) {\makebox[0.54cm][c]{\textcolor[HTML]{FFFFFF}{\tiny 0.9}}};}} & \colorbox[HTML]{003A8A}{\makebox[0.54cm][c]{\textcolor[HTML]{FFFFFF}{\tiny 0.7}}} & \colorbox[HTML]{003A8A}{\makebox[0.54cm][c]{\textcolor[HTML]{FFFFFF}{\tiny 0.6}}} & {\definecolor{ldbg}{HTML}{0C519B}%
\tikz[baseline=(ldc.base)]{\node[inner sep=1.2pt,outer sep=0pt,line width=0.6pt,draw=black,dash pattern=on 1.1pt off 0.9pt,fill=ldbg] (ldc) {\makebox[0.44cm][c]{\textcolor[HTML]{FFFFFF}{\tiny 0.3}}};}} & {\definecolor{ldbg}{HTML}{003A8A}%
\tikz[baseline=(ldc.base)]{\node[inner sep=1.2pt,outer sep=0pt,line width=0.6pt,draw=black,dash pattern=on 1.1pt off 0.9pt,fill=ldbg] (ldc) {\makebox[0.44cm][c]{\textcolor[HTML]{FFFFFF}{\tiny 0.3}}};}} & {\definecolor{ldbg}{HTML}{003A8A}%
\tikz[baseline=(ldc.base)]{\node[inner sep=1.2pt,outer sep=0pt,line width=0.6pt,draw=black,dash pattern=on 1.1pt off 0.9pt,fill=ldbg] (ldc) {\makebox[0.44cm][c]{\textcolor[HTML]{FFFFFF}{\tiny 0.3}}};}} & \colorbox[HTML]{003A8A}{\makebox[0.44cm][c]{\textcolor[HTML]{FFFFFF}{\tiny 0.2}}} \\
\end{tabular}
\end{table*}

\begin{figure*}[ht!]
\centering
\includegraphics[width=\linewidth]{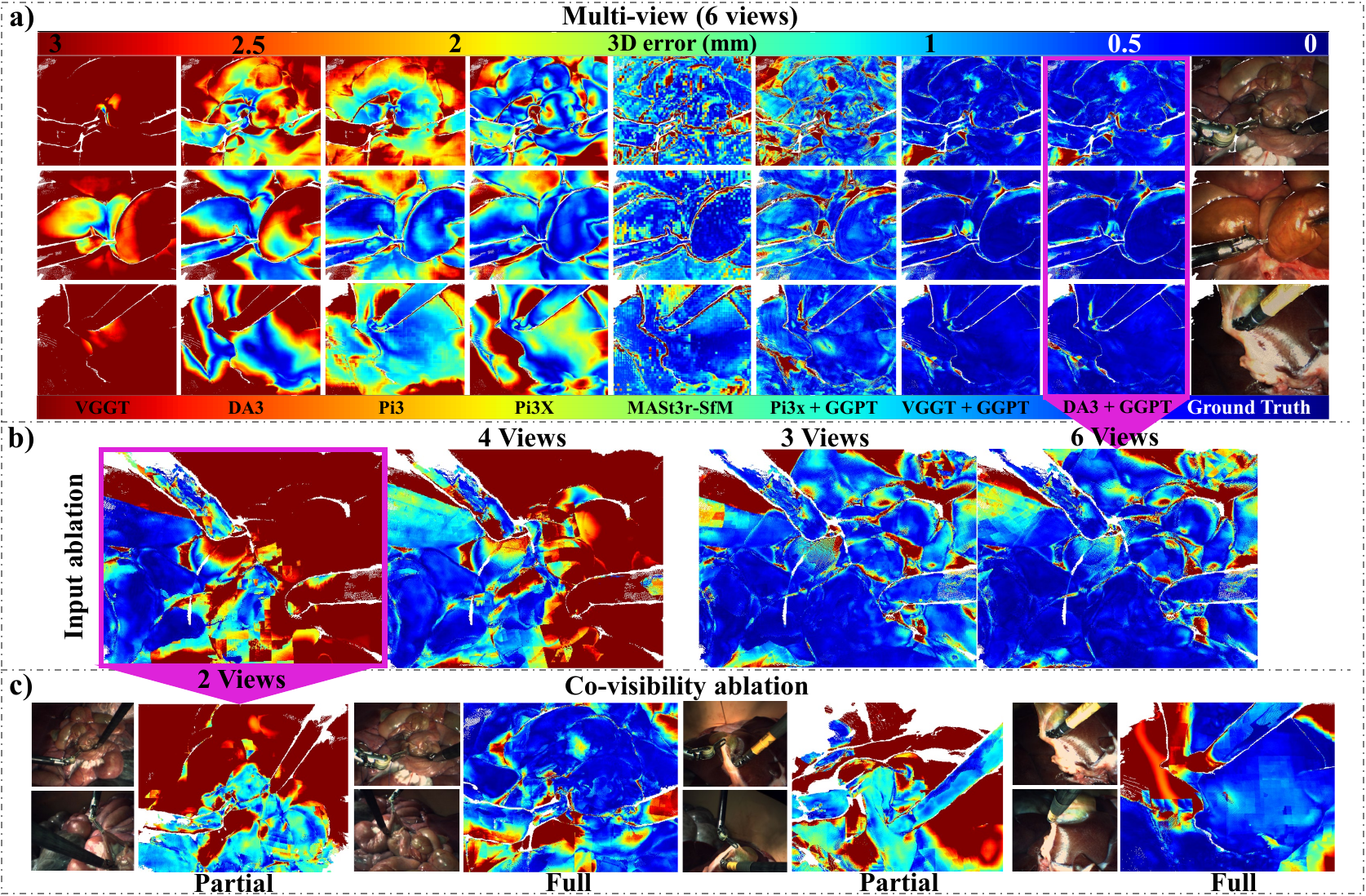}
\caption{Representative visual results for multi-view methods on the MV-dVRK benchmark. 
(a) Mean error (ME) in the reference view E1L across evaluated multi-view methods at six synchronized views; geometry-refined methods achieve the lowest and most spatially uniform errors. 
(b) For Depth Anything~3~\cite{lin2025depth} + GGPT~\cite{chen2026ggpt}, increasing input viewpoint count progressively reduces reconstruction error. 
(c) Co-visibility analysis at the sparsest setting: partial view overlap leaves large regions without cross-view geometric support.}

\label{fig:multiview_benchmark}
\end{figure*}

\subsection{Implementation details for selected methods}
\label{sec:implementation}

We use publicly released implementations and default inference settings unless stated otherwise. Endo3R~\cite{guo2025endo3r} is evaluated with two-frame and 50-frame temporal windows and uses fitted Sim(3) registration. FoundationStereo~\cite{wen2025foundationstereo} uses calibrated stereo rectification and is evaluated with both fitted and camera-anchored alignment. For multi-stereo, independently reconstructed FoundationStereo point clouds are registered using RoMa\,v2~\cite{edstedt2025roma} correspondences with either rigid SE(3) or similarity Sim(3) transforms. Multi-view methods use their standard pipelines, with calibrated intrinsics supplied where supported. The supplementary material provides further preprocessing and registration details.

\subsection{Quantitative results}
\label{sec:results:quantitative}

Tab.~\ref{tab:results_approaches_compared} and Fig.~\ref{fig:registration_offset} summarize the main comparisons across reconstruction paradigms. Tab.~\ref{tab:results_fused_split} provides the full comparison of multi-view methods, while Fig.~\ref{fig:multiview_benchmark} visualizes their geometric errors and the effects of input viewpoint count.

\textbf{Single-view reconstruction.}
Dense stereo provides substantially greater scene coverage than monocular reconstruction from the same endoscopic viewpoint. Compared with the two-frame Endo3R baseline, FoundationStereo increases coverage by 27 percentage points (95\% CI [20, 34], 8/8 scenes). Temporal accumulation narrows this gap but does not close it: even after 50 monocular frames, stereo retains a 15-point coverage advantage (95\% CI [5.8, 25], 6/8). This finding indicates that temporal observations from one moving camera do not recover all of the geometry available from simultaneous stereo observations.

\textbf{Multi-stereo reconstruction.}
Additional endoscopes increase coverage by exposing anatomy that is occluded from the reference viewpoint (Fig.~\ref{fig:registration_offset}d). Registering a second stereo reconstruction increases coverage by 14 percentage points (95\% CI [7.2, 21], 8/8), and a third viewpoint provides a further 7.2-point gain (95\% CI [5.2, 9.3], 8/8). The local geometry of each stereo reconstruction remains accurate, but cross-endoscope registration can introduce duplicated structures and other inconsistencies (Fig.~\ref{fig:registration_offset}e). Registration quality depends strongly on correspondence selection and solver robustness; the corresponding ablations are provided in the supplementary material.

\textbf{Multi-view reconstruction.}
Unlike multi-stereo, multi-view methods jointly reconstruct all input images in a common geometric frame and can therefore use additional viewpoints to improve the reconstruction itself. Adding the third viewpoint consistently reduces error in the common reference view E1L. For geometry-refined methods, the mean reduction is 1.9\,mm (95\% CI [1.2, 2.6], 8/8 scenes), corresponding to a 54\% relative improvement. The effect on E2L is more variable: the mean reduction is 2.1\,mm (95\% CI [$-$0.05, 4.2]), reflecting differences in inter-endoscope geometry and co-visibility across scenes.

These results highlight a clear crossover between reconstruction paradigms. With two endoscopes, registered stereo is more accurate in E1L than every evaluated multi-view method, with the advantage holding across all eight scenes. With three endoscopes, geometry-refined multi-view reconstruction overtakes multi-stereo in the reference view and achieves the highest overall coverage. At six synchronized views, the refined methods reach 66--67\% coverage within 1\,mm, compared with 28--44\% for the feed-forward models (Tab.~\ref{tab:results_fused_split}). Thus, the benefit of explicit geometric optimization becomes most pronounced once sufficient cross-view support is available.

\textbf{Role of view overlap.}
The remaining failures are strongly linked to sparse geometric support. At comparable working distance, reduced co-visibility between viewpoints is associated with increased reconstruction error for most multi-view methods; analysis is provided in the supplementary material. Fig.~\ref{fig:multiview_benchmark}c illustrates this behavior for Depth Anything~3~\cite{lin2025depth} + GGPT~\cite{chen2026ggpt}: under partial overlap, large regions receive no cross-view geometric constraints, forcing the point transformer to rely more on its learned prior and producing inaccurate 3D predictions.

\subsection{Implications for multi-view surgical reconstruction}
\label{sec:discussion}

Our results suggest complementary roles for calibrated stereo and global multi-view optimization. Stereo remains particularly effective in the sparsest regime, while geometry-refined multi-view methods benefit strongly from additional cross-view support. A promising direction is therefore rig-aware optimization that preserves the known geometry within each stereo endoscope while jointly refining scene structure and relative poses across endoscopes. Practical deployment will additionally require lower-latency dense matching and optimization. MV-dVRK II extends these questions to dynamic scenes involving camera motion, tissue deformation, and robot kinematics.

\subsection{MV-dVRK limitations}
\label{sec:limitations}

MV-dVRK I provides dense reference geometry only for static tissue; obtaining comparable ground truth under deformation remains impractical, so MV-dVRK II does not provide dense surface geometry. The camera poses estimated with dense SfM also cannot be externally validated without introducing manual and optical calibration biases.
The static subset of the benchmark contains eight scenes from two porcine specimens and three surgical tasks. This dataset supports rigorous paired comparisons across methods within the benchmark, but the measured effect sizes should not be assumed to transfer unchanged to new anatomies, procedures, or in-vivo conditions. Finally, the reference surface is reconstructed rather than directly measured, though it is independently validated using an industrial 3D scanner, and the open ex-vivo acquisition with additional illumination differs from a closed abdominal cavity.

\section{Conclusion}
\label{sec:conclusion}

We introduced MV-dVRK, a surgical dataset combining three exposure-synchronized stereo viewpoints with validated reference geometry and camera poses, and we used it to establish a controlled benchmark for sparse multi-view 3D reconstruction on real robotic surgery data. Our results reveal a clear transition with viewpoint count: registered stereo is highly competitive in the sparsest two-endoscope setting, while geometry-refined multi-view methods benefit strongly from a third viewpoint and achieve the best coverage and reference-view accuracy. Beyond the static benchmark, synchronized dynamic sequences extend MV-dVRK to camera motion and tissue deformation, providing a basis for future multi-viewpoint surgical 3D perception.

\section{Acknowledgments}
\label{sec:ackn}
The authors thank the International Max Planck Research School for Intelligent Systems (IMPRS-IS) and the Max Planck ETH Center for Learning Systems (CLS) for supporting Guido Caccianiga.\\

{
    \small
    \bibliographystyle{ieeenat_fullname}
    \bibliography{references}

@misc{ieee1588,
  title        = {{IEEE} Standard for a Precision Clock Synchronization Protocol for Networked Measurement and Control Systems},
  author       = {{IEEE}},
  howpublished = {IEEE Std 1588-2019},
  year         = {2020},
  doi          = {10.1109/IEEESTD.2020.9120376}
}

@inproceedings{LORANSAC_DBLP:conf/dagm/ChumMK03,
  author       = {Ondrej Chum and
                  Jiri Matas and
                  Josef Kittler},
  editor       = {Bernd Michaelis and
                  Gerald Krell},
  title        = {Locally Optimized {RANSAC}},
  booktitle    = {Joint Pattern Recognition Symposium},
  series       = {Lecture Notes in Computer Science},
  volume       = {2781},
  pages        = {236--243},
  publisher    = {Springer},
  year         = {2003},
  url          = {https://doi.org/10.1007/978-3-540-45243-0\_31},
  doi          = {10.1007/978-3-540-45243-0\_31},
  bibsource    = {dblp computer science bibliography, https://dblp.org}
}

@ARTICLE{Umeyama88573,
  author={Umeyama, S.},
  journal={IEEE Transactions on Pattern Analysis and Machine Intelligence}, 
  title={Least-squares estimation of transformation parameters between two point patterns}, 
  year={1991},
  volume={13},
  number={4},
  pages={376-380},

  doi={10.1109/34.88573}}

@inproceedings{inproceedings,
author = {Zinßer, Timo and Schmidt, Jochen and Niemann, Heinrich},
year = {2005},
month = {01},
title = {Point set registration with integrated scale estimation},
booktitle = {International Conference on Pattern Recognition and Information Processing (PRIP)}
}

@misc{arteceva,
  author       = {{Artec 3D}},
  title        = {Artec {Eva} {3D} Scanner},
  howpublished = {\url{https://www.artec3d.com/portable-3d-scanners/artec-eva}},
  note         = {Accessed: 28 August 2026},
  year = 2026,
}

@book{hartley2003mvg,
  author    = {Hartley, R. and Zisserman, A.},
  title     = {Multiple View Geometry in Computer Vision},
  publisher = {Cambridge University Press},
  edition   = {2nd},
  year      = {2003},
  isbn      = {9780521540513}
}

@inproceedings{schoenberger2016sfm,
  author    = {J. L. Sch{\"o}nberger and J.-M. Frahm},
  title     = {Structure-from-Motion Revisited},
  booktitle = {IEEE/CVF Conference on Computer Vision and Pattern Recognition (CVPR)},
  year      = {2016},
  pages     = {4104--4113},
  doi       = {10.1109/CVPR.2016.445}
}

@inproceedings{wang2024vggsfm,
  title={{VGGSfM:} Visual Geometry Grounded Deep Structure From Motion},
  author={Wang, Jianyuan and Karaev, Nikita and Rupprecht, Christian and Novotny, David},
  booktitle={IEEE/CVF Conference on Computer Vision and Pattern Recognition (CVPR)},
  pages={21686--21697},
  year={2024}
}

@inproceedings{wang2022neural,
  title={Neural rendering for stereo {3D} reconstruction of deformable tissues in robotic surgery},
  author={Wang, Yuehao and Long, Yonghao and Fan, Siu Hin and Dou, Qi},
  booktitle={International Conference on Medical Image Computing and Computer-Assisted Intervention (MICCAI)},
  pages={431--441},
  month=sep,
  year={2022},
  doi={10.1007/978-3-031-16449-1_41},
  organization={Springer}
}

@inproceedings{moccia2019vision,
author = {Moccia, Rocco and Selvaggio, Mario and Villani, Luigi and Siciliano, Bruno and Ficuciello, Fanny},
title = {Vision-based virtual fixtures generation for robotic-assisted polyp dissection procedures},
year = {2019},
doi = {10.1109/IROS40897.2019.8968080},
booktitle = {IEEE/RSJ International Conference on Intelligent Robots and Systems (IROS)},
pages = {7934--7939},
numpages = {6},
}

@article{li2020super,
  author={Li, Yang and Richter, Florian and Lu, Jingpei and Funk, Emily K. and Orosco, Ryan K. and Zhu, Jianke and Yip, Michael C.},
  journal={IEEE Robotics and Automation Letters}, 
  title={{SuPer:} A surgical perception framework for endoscopic tissue manipulation with surgical robotics}, 
  year={2020},
  volume={5},
  number={2},
  pages={2294--2301},
  doi={10.1109/LRA.2020.2970659}
}

@article{boal2024evaluation,
  title={Evaluation status of current and emerging minimally invasive robotic surgical platforms},
  author={Boal, Matthew and Di Girasole, C. Giovene and Tesfai, Freweini and Morrison, T. E. M. and Higgs, Simon and Ahmad, Jawad and Arezzo, Alberto and Francis, Nader},
  journal={Surgical Endoscopy},
  volume={38},
  number={2},
  pages={554--585},
  year={2024},
  doi={10.1007/s00464-023-10554-4},
  organization={Springer}
}

@article{song2026diff2dgs,
  author = {Song, Tianyi and Stoyanov, Danail and Mazomenos, Evangelos and Vasconcelos, Francisco},
  title = {{Diff2DGS:} Reliable reconstruction of occluded surgical scenes via {2D Gaussian} splatting},
  journal = {IEEE Robotics and Automation Letters},
  year=2026,
  volume={11},
  number={10},
  pages={11251--11258},
  doi={10.1109/LRA.2026.3723286}
}

@article{yu2026endopairgs,
  title = {{Endo-PairGS:} Pair priors for dynamic endoscopic scene reconstruction},
  author = {Yu, Xiankang and Hayashi, Yuichiro and Oda, Masahiro and Kitasaka, Takayuki and Mori, Kensaku},
  journal = {International Journal of Computer Assisted Radiology and Surgery},
  volume = {21},
  pages = {1341--1350},
  year = {2026},
  doi = {10.1007/s11548-026-03707-y}
}

@inproceedings{zhu2024deformable,
  title={{EndoGS:} Deformable endoscopic tissues reconstruction with {G}aussian splatting},
  author={Zhu, Lingting and Wang, Zhao and Jin, Zhenchao and Lin, Guying and Yu, Lequan},
  booktitle={International Conference on Medical Image Computing and Computer-Assisted Intervention (MICCAI)},
  pages={135--145},
  month=oct,
  year={2024},
  doi={10.1007/978-3-031-77610-6_13},
  organization={Springer}
}

@misc{moghani2025sufia,
	title = {{SuFIA-BC}: Generating high quality demonstration data for visuomotor policy learning in surgical subtasks},
	author = {Moghani, Masoud and Nelson, Nigel and Ghanem, Mohamed and Diaz-Pinto, Andres and Hari, Kush and Azizian, Mahdi and Goldberg, Ken and Huver, Sean and Garg, Animesh},
	year = {2025},
	note = {\textit{arXiv:2504.14857}},
    doi = {10.48550/arXiv.2504.14857},
}

@inproceedings{kazanzides2014open,
  title={An open-source research kit for the da {Vinci{\textregistered} Surgical System}},
  author={Kazanzides, Peter and Chen, Zihan and Deguet, Anton and Fischer, Gregory S. and Taylor, Russell H. and DiMaio, Simon P.},
  booktitle={IEEE International Conference on Robotics and Automation (ICRA)},
  pages={6434--6439},
  year={2014},
  doi={10.1109/ICRA.2014.6907809}
}

@misc{allan2021stereo,
  title={Stereo correspondence and reconstruction of endoscopic data challenge},
  author={Allan, Max and Mcleod, Jonathan and Wang, Congcong and Rosenthal, Jean Claude and Hu, Zhenglei and Gard, Niklas and Eisert, Peter and Fu, Ke Xue and Zeffiro, Trevor and Xia, Wenyao and Zhu, Z. and Luo, H. and Zhang, X. and Li, X. and Jia, F. and Sharan, L. and Kurmann, T. and Schmid, S. and Sznitman, R. and Psychogyios, D. and Azizian, M. and Stoyanov, D. and Maier-Hein, L. and Speidel, S.},
  note={\textit{arXiv:2101.01133}},
  doi={10.48550/arXiv.2101.01133},
  month=jan,
  year={2021}
}

@article{edwards2022serv,
  title={{SERV-CT:} A disparity dataset from cone-beam {CT} for validation of endoscopic {3D} reconstruction},
  author={Edwards, P. J. Eddie and Psychogyios, Dimitris and Speidel, Stefanie and Maier-Hein, Lena and Stoyanov, Danail},
  journal={Medical Image Analysis},
  volume={76},
  note={Art.\ no.\ 102302},
  year={2022},
  doi={0.1016/j.media.2021.102302},
  organization={Elsevier}
}

@article{penza2018endoabs,
  title={{EndoAbS dataset:} Endoscopic abdominal stereo image dataset for benchmarking {3D} stereo reconstruction algorithms},
  author={Penza, Veronica and Ciullo, Andrea S. and Moccia, Sara and Mattos, Leonardo S. and De Momi, Elena},
  journal={The International Journal of Medical Robotics and Computer Assisted Surgery},
  volume={14},
  number={5},
  note={Art.\ no.\ e1926},
  year={2018},
  doi={10.1002/rcs.1926},
  organization={Wiley Online Library}
}

@article{hayoz2023learning,
  title={Learning how to robustly estimate camera pose in endoscopic videos},
  author={Hayoz, Michel and Hahne, Christopher and Gallardo, Mathias and Candinas, Daniel and Kurmann, Thomas and Allan, Maximilian and Sznitman, Raphael},
  journal={International Journal of Computer Assisted Radiology and Surgery},
  volume={18},
  number={7},
  pages={1185--1192},
  year={2023},
  doi={10.1007/s11548-023-02919-w},
  organization={Springer}
}

@article{jiang2025surgisr4k,
  title={{SurgiSR4K:} A high-resolution endoscopic video dataset for robotic-assisted minimally invasive procedures},
  author={Jiang, Fengyi and Zhang, Xiaorui and Jin, Lingbo and Liang, Ruixing and Chen, Yuxin and Venkatesh, Adi Chola and Culman, Jason and Wu, Tiantian and Shao, Lirong and Sun, Wenqing and Gao, Cong and McNamara, Hallie and Lu, Jingpei and Mohareri, Omid},
  journal={Machine Learning for Biomedical Imaging},
  volume={3},
  pages={875--885},
  doi={10.59275/j.melba.2025-f593},
  year={2025}
}

@misc{zia2025surgical,
  title={Surgical visual understanding {(SurgVU)} dataset},
  author={Zia, Aneeq and Berniker, Max and Nespolo, Rogerio and Perreault, Conor and Wang, Ziheng and Mueller, Benjamin and Schmidt, Ryan and Bhattacharyya, Kiran and Liu, Xi and Jarc, Anthony},
  note={\textit{arXiv:2501.09209}},
  doi={10.48550/arXiv.2501.09209},
  month=may,
  year={2026}
}

@article{hansen2026imitatecholec,
  author = {Hansen, Pascal and Kim, Ji Woong Brian and Goldenberg, Antony and Chen, Juo Tung and Li, Yuanzhe Amos and Deguet, Anton and White, Brandon and Tsai, De Ru and Cha, Richard and Jopling, Jeffrey and Scheikl, Paul Maria and Krieger, Axel},
  title = {{ImitateCholec:} A multimodal dataset for long-horizon imitation learning in robotic cholecystectomy},
  journal = {Scientific Data},
  volume = {13},
  year = {2026},
  doi = {10.1038/s41597-025-06526-z},
  note = {{Art.\ no.\ 215}},
  organization={Nature Publishing Group UK London}
}

@inproceedings{bonilla2026imed,
  title = {{iMED:} A multi-endoscope dataset for surgical {3D} perception},
  author = {Bonilla, Sierra and Jiang, Fengyi and Nwoye, Chinedu and Lu, Jingpei and Reardon, Kailey and Chow, Humphrey W. and Vasconcelos, Francisco and Bano, Sophia and Schmidt, Adam and Mohareri, Omid},
  booktitle = {European Conference on Computer Vision (ECCV)},
  year = {2026},
  note = {In press}
}

@article{garrido-jurado2014automatic,
title = {Automatic generation and detection of highly reliable fiducial markers under occlusion},
author = {Garrido-Jurado, S. and Muñoz-Salinas, R. and Madrid-Cuevas, F.J. and Marín-Jiménez, M.J.},
journal = {Pattern Recognition},
volume = {47},
number = {6},
pages = {2280-2292},
year = {2014},
doi = {10.1016/j.patcog.2014.01.005},
}

@inproceedings{jing2023uncertainty,
  title={Uncertainty guided adaptive warping for robust and efficient stereo matching},
  author={Jing, Junpeng and Li, Jiankun and Xiong, Pengfei and Liu, Jiangyu and Liu, Shuaicheng and Guo, Yichen and Deng, Xin and Xu, Mai and Jiang, Lai and Sigal, Leonid},
  booktitle={IEEE/CVF International Conference on Computer Vision (ICCV)},
  pages={3295--3304},
  month=oct,
  year={2023},
  doi={10.1109/ICCV51070.2023.00307},
}

@inproceedings{lipson2021raft,
  title={{RAFT-Stereo:} Multilevel recurrent field transforms for stereo matching},
  author={Lipson, Lahav and Teed, Zachary and Deng, Jia},
  booktitle={International Conference on 3D Vision (3DV)},
  pages={218--227},
  month=dec,
  year={2021},
  organization={IEEE},
  doi={10.1109/3DV53792.2021.00032}
}

@inproceedings{wang2024selective,
  title={{Selective-Stereo:} Adaptive frequency information selection for stereo matching},
  author={Wang, Xianqi and Xu, Gangwei and Jia, Hao and Yang, Xin},
  booktitle={IEEE/CVF Conference on Compututer Vision and Pattern Recognition (CVPR)},
  pages={19701--19710},
  month=jun,
  year={2024}
}

@inproceedings{wen2025foundationstereo,
  title={{FoundationStereo:} Zero-shot stereo matching},
  author={Wen, Bowen and Trepte, Matthew and Aribido, Joseph and Kautz, Jan and Gallo, Orazio and Birchfield, Stan},
  booktitle={IEEE/CVF Conference on Computer Vision and Pattern Recognition (CVPR)},
  pages={5249--5260},
  doi={10.1109/CVPR52734.2025.00495},
  city={Nashville, USA},
  day={10--17},
  month=jun,
  year={2025}
}

@inproceedings{guo2025endo3r,
  title={{Endo3R:} Unified online reconstruction from dynamic monocular endoscopic video},
  author={Guo, Jiaxin and Dong, Wenzhen and Huang, Tianyu and Ding, Hao and Wang, Ziyi and Kuang, Haomin and Dou, Qi and Liu, Yun-Hui},
  booktitle={International Conference on Medical Image Computing and Computer-Assisted Intervention (MICCAI)},
  pages={170--180},
  year={2025},
  doi={10.1007/978-3-032-05114-1_17},
  organization={Springer}
}

@inproceedings{sheikh2025endo,
  title={{Endo-FASt3r:} Endoscopic foundation model adaptation for structure from motion},
  author={Sheikh Zeinoddin, Mona and Hoque, Mobarak I. and Tandogdu, Zafer and Shaw, Greg L. and Clarkson, Matthew J. and Mazomenos, Evangelos B. and Stoyanov, Danail},
  booktitle={International Conference on Medical Image Computing and Computer-Assisted Intervention (MICCAI)},
  pages={117--126},
  year={2025},
  organization={Springer}
}

@article{schmidt2024tracking,
title = {Tracking and mapping in medical computer vision: A review},
author = {Schmidt, Adam and Mohareri, Omid and DiMaio, Simon and Yip, Michael C. and Salcudean, Septimiu E.},
journal = {Medical Image Analysis},
volume = {94},
note = {Art.\ No.\ 103131},
year = {2024},
doi = {10.1016/j.media.2024.103131},
}

@inproceedings{tretschk2023state,
  title={State of the art in dense monocular non-rigid {3D} reconstruction},
  author={Tretschk, Edith and Kairanda, Navami and B R, Mallikarjun and Dabral, Rishabh and Kortylewski, Adam and Egger, Bernhard and Habermann, Marc and Fua, Pascal and Theobalt, Christian and Golyanik, Vladislav},
  booktitle={Computer Graphics Forum},
  volume={42},
  number={2},
  pages={485--520},
  year={2023},
  organization={Wiley Online Library},
  doi={10.1111/cgf.14774}
}

@inproceedings{wang2024dust3r,
  title={{DUSt3R}: Geometric {3D} vision made easy},
  author={Wang, Shuzhe and Leroy, Vincent and Cabon, Yohann and Chidlovskii, Boris and Revaud, Jerome},
  booktitle={IEEE/CVF Conference on Computer Vision and Pattern Recognition (CVPR)},
  pages={20697--20709},
  month={June},
  year={2024},
  doi={10.48550/arXiv.2312.14132}
}

@misc{lin2025depth,
  title={{Depth Anything 3:} Recovering the visual space from any views},
  author={Lin, Haotong and Chen, Sili and Liew, Junhao and Chen, Donny Y. and Li, Zhenyu and Shi, Guang and Feng, Jiashi and Kang, Bingyi},
  doi={10.48550/arXiv.2507.01634},
  note={\textit{arXiv:2511.10647}},
  month=jul,
  year={2025}
}

@inproceedings{stoyanov2010realtime,
  author = {Stoyanov, Danail and Visentini-Scarzanella, Marco
  and Pratt, Philip and Yang, Guang-Zhong},
  title = {Real-time stereo reconstruction in robotically assisted minimally invasive surgery},
  booktitle = {International Conference on Medical Image Computing and Computer-Assisted Intervention (MICCAI)},
  pages = {275--282},
  year = {2010},
  organization = {Springer},
  doi = {10.1007/978-3-642-15705-9_34}
}

@inproceedings{cheng2022deep,
  author = {Cheng, Xuelian and Zhong, Yiran and Harandi, Mehrtash and Drummond, Tom and Wang, Zhiyong and Ge, Zongyuan},
  title = {Deep laparoscopic stereo matching with transformers},
  booktitle = {International Conference on Medical Image Computing and Computer-Assisted Intervention (MICCAI)},
  pages = {464--474},
  year = {2022},
  organization = {Springer},
  doi = {10.1007/978-3-031-16449-1_44}
}

@article{mahmoud2019live,
  author = {Mahmoud, Nader and Collins, Toby and Hostettler, Alexandre and Soler, Luc and Doignon, Christophe and Montiel, Jose Maria Martinez},
  title = {Live tracking and dense reconstruction for handheld monocular endoscopy},
  journal = {IEEE Transactions on Medical Imaging},
  volume  = {38},
  number  = {1},
  pages   = {79--89},
  year    = {2019},
  doi     = {10.1109/TMI.2018.2856109}
}

@misc{hamlyn,
  title = {Hamlyn {C}entre surgical dataset},
  howpublished = {\url{http://hamlyn.doc.ic.ac.uk/vision/}},
  note = {Accessed: 4 June 2023},
  year={2010},
  key = {{Hamlyn Centre}},
}

@inproceedings{chen2026ggpt,
  title={{GGPT}: Geometry-grounded point transformer},
  author={Chen, Yutong and Wang, Yiming and Zhang, Xucong and Prokudin, Sergey and Tang, Siyu},
  booktitle={IEEE/CVF Conference on Computer Vision and Pattern Recognition (CVPR)},
  pages={28959--28968},
  doi={10.48550/arXiv.2603.11174},
  month=jun,
  year={2026},
}

@inproceedings{duisterhof2025mast3r,
  title={{MASt3R-SfM:} A fully-integrated solution for unconstrained structure-from-motion},
  author={Duisterhof, Bardienus Pieter and Zust, Lojze and Weinzaepfel, Philippe and Leroy, Vincent and Cabon, Yohann and Revaud, Jerome},
  booktitle={International Conference on 3D Vision (3DV)},
  pages={1--10},
  month=mar,
  year={2025},
  doi={10.1109/3DV66043.2025.00008},
  organization={IEEE}
}

@inproceedings{leroy2024mast3r,
  author    = {Vincent Leroy and Yohann Cabon and Jerome Revaud},
  title     = {Grounding Image Matching in {3D} with {MASt3R}},
  booktitle = {European Conference on Computer Vision (ECCV)},
  pages={71--91},
  year      = {2024},
  doi       = {10.1007/978-3-031-73220-1_5}
}

@inproceedings{wang2025vggt,
  title={{VGGT:} Visual geometry grounded transformer},
  author={Wang, Jianyuan and Chen, Minghao and Karaev, Nikita and Vedaldi, Andrea and Rupprecht, Christian and Novotny, David},
  booktitle={IEEE/CVF Conference on Computer Vision and Pattern Recognition (CVPR)},
  pages={5294--5306},
  month={June},
  year={2025},
  doi={10.48550/arXiv.2503.11651}
}

@inproceedings{wang2025pi,
  title={{$\pi^{3}$: Permutation-equivariant} visual geometry learning},
  author={Wang, Yifan and Zhou, Jianjun and Zhu, Haoyi and Chang, Wenzheng and Zhou, Yang and Li, Zizun and Chen, Junyi and Pang, Jiangmiao and Shen, Chunhua and He, Tong},
  booktitle={International Conference on Learning Representations (ICLR)},
  year={2025},
  doi={10.48550/arXiv.2507.13347}
}

@misc{wang2025pi3x,
  title        = {{Pi3X} model update for ``{P}ermutation-equivariant visual geometry learning''},
  author       = {Wang, Yifan and Zhou, Jianjun and Zhu, Haoyi and Chang, Wenzheng and Zhou, Yang and Li, Zizun and Chen, Junyi and Pang, Jiangmiao and Shen, Chunhua and He, Tong},
  howpublished = {GitHub Repository},
  month        = dec,
  year         = {2025},
  note         = {Accessed: 1 July 2026},
  url          = {https://github.com/yyfz/Pi3}
}

@misc{edstedt2025roma,
  title={{RoMa v2:} Harder better faster denser feature matching},
  author={Edstedt, Johan and Nordstr{\"o}m, David and Zhang, Yushan and B{\"o}kman, Georg and Astermark, Jonathan and Larsson, Viktor and Heyden, Anders and Kahl, Fredrik and Wadenb{\"a}ck, M{\aa}rten and Felsberg, Michael},
  note={\textit{arXiv: 2511.15706}, Accepted to ECCV 2026},
  doi={10.48550/arXiv.2511.15706},
  year={2026}
}

@misc{caccianiga2024dense,
  title={Dense {3D} reconstruction through lidar: A comparative study on ex-vivo porcine tissue},
  author={Caccianiga, Guido and Nubert, Julian and Hutter, Marco and Kuchenbecker, Katherine J.},
  year={2024},
  note = {\textit{arXiv:2401.10709}},
  doi = {10.48550/arXiv.2401.10709}
}
}

\end{document}